\pdfoutput=1
\documentclass[letterpaper]{article}
\usepackage[colorlinks=true,citecolor=blue]{hyperref}
\usepackage{uai2020}
\usepackage[margin=1in]{geometry}

\usepackage{times}
\usepackage{format}
\usepackage{placeins}

\title{Mutual Information Based Knowledge Transfer Under State-Action Dimension Mismatch}

\author{} 

\author{ {\bf Michael Wan} \\
Computer Science Dept. \\
UIUC\\
mw3@illinois.edu \\
\And
{\bf Tanmay Gangwani}  \\
Computer Science Dept. \\
UIUC         \\
gangwan2@illinois.edu \\
\And
{\bf Jian Peng}   \\
Computer Science Dept. \\
UIUC    \\
jianpeng@illinois.edu \\
}

\begin{document}

\maketitle

\begin{abstract}
Deep reinforcement learning (RL) algorithms have achieved great success on a wide variety of sequential decision-making tasks. However, many of these algorithms suffer from high sample complexity when learning from scratch using environmental rewards, due to issues such as credit-assignment and high-variance gradients, among others. Transfer learning, in which knowledge gained on a source task is applied to more efficiently learn a different but related target task, is a promising approach to improve the sample complexity in RL. Prior work has considered using pre-trained teacher policies to enhance the learning of the student policy, albeit with the constraint that the teacher and the student MDPs share the state-space or the action-space. In this paper, we propose a new framework for transfer learning where the teacher and the student can have arbitrarily different state- and action-spaces. To handle this mismatch, we produce embeddings which can systematically extract knowledge from the teacher policy and value networks, and blend it into the student networks. To train the embeddings, we use a task-aligned loss and show that the representations could be enriched further by adding a mutual information loss. Using a set of challenging simulated robotic locomotion tasks involving many-legged centipedes, we demonstrate successful transfer learning in situations when the teacher and student have different state- and action-spaces.

\end{abstract}

\section{INTRODUCTION}
Deep reinforcement learning (RL), which combines the rigor of RL algorithms with the flexibility of universal function approximators such as deep neural networks, has demonstrated a plethora of success stories in recent times. These include computer and board games~\citep{mnih2015human,silver2016mastering}, continuous control~\citep{lillicrap2015continuous}, and robotics~\citep{rajeswaran2017learning}, to name a few. Crucially though, these methods have been shown to be performant in the regime where an agent can accumulate vast amounts of experience in the environment, usually modeled with a simulator. For real-world environments such as autonomous navigation and industrial processes, data generation is an expensive (and sometimes risky) procedure. To make deep RL algorithms more sample-efficient, there is great interest in designing techniques for {\em knowledge transfer}, which enables accelerating agent learning by leveraging either existing trained policies (referred to as teachers), or using task demonstrations for imitation learning~\citep{abbeel2004apprenticeship}. One promising idea for knowledge transfer in RL is policy distillation~\citep{rusu2015policy, parisotto2015actor, hinton2015distilling}, where information from the teacher policy network is transferred to a student policy network to improve the learning process. 

Prior work has incorporated policy distillation in a variety of settings~\citep{czarnecki2019distilling}. Some examples include the transfer of knowledge from simple to complex agents while following a curriculum over agents~\citep{czarnecki2018mix}, learning a centralized policy that captures shared behavior across tasks for multi-task RL~\citep{teh2017distral}, distilling information from parent policies into a child policy for a genetically-inspired RL algorithm~\citep{gangwani2017policy}, and speeding-up large-scale population-based training using multiple teachers~\citep{schmitt2018kickstarting}. A common motif in these approaches is the use of Kullback-Leibler (KL) divergence between the state-conditional action distributions of the teacher and student networks, as the minimization objective for knowledge transfer. While simple and intuitive, this restricts learning from teachers that have the same output (action) space as the student, since KL divergence is only defined for distribution over a common space. An alternative to knowledge sharing in the action-space is information transfer through the embedding-space formed via the different layers of a deep neural network.~\citep{liu2019knowledge} provides an example of this; it utilizes learned lateral connections between intermediate layers of the teacher and student networks. Although the action-spaces can now be different, the state-space is still required to be identical between the teacher and the student, since the same input observation is fed to both the networks~\citep{liu2019knowledge}. 

In our work, we present a transfer learning approach to accelerate the training of the student policy, by leveraging teacher policies trained in an environment with {\em different state- and action-space}. Arguably, there is a huge potential for data-efficient student learning by tapping into teachers trained on dissimilar, but related tasks. For instance, consider an available teacher policy for locomotion of a quadruped robot, where the (97-dimensional) state-space is the set of joint-angles and joint-velocities and the (10-dimensional) action-space is the torques to the joints. If we wish to learn locomotion for a hexapod robot (state-dimension 139, action-dimension 16), we conjecture that the learning could be kick-started by harnessing the information stored in the trained neural network for the quadruped, since both the tasks are locomotion for legged robots and therefore share an inherent structure. However, the dissimilar state- and action-space preclude the use of the knowledge transfer mechanisms proposed in prior work. 

Our approach deals with the mismatch in the state- and action-space of the teacher and student in the following manner. To handle disparate actions, rather than using divergence minimization in the action-space, we transfer knowledge by augmenting representations in the layers of the student network with representations from the layers of the teacher network. This is similar to the knowledge flow in~\citep{liu2019knowledge} using lateral connections, but with the important difference that we do not employ learnable matrices to transform the teacher representation. The mismatch in the observation- or state-space has not been considered in prior literature, to the best of our knowledge. We manage this by learning an embedding space which can be used to extract the necessary information from the available teacher policy network. These embeddings are trained to adhere to two properties. Firstly, they must be task-aligned. Our RL objective is the maximization of cumulative discounted rewards in the student environment, and therefore, the embeddings must be aligned to serve that goal. Secondly, we would like the embeddings to be correlated with the states encountered by the student policy. The embeddings are used to deterministically draw out knowledge from the teacher network. Therefore, a high correlation ensures that the most suitable teacher guidance is derived for each student state. We achieve this by maximizing the mutual information between the embeddings and student states. We evaluate our method on a set of challenging robotic locomotion tasks modeled using the MuJoCo simulator. We demonstrate the successful transfer of knowledge from trained teachers to students, in the scenario of mismatched state- and action-space. This leads to appreciable gains in sample-efficiency, compared to RL from scratch using only the environmental rewards.
\label{sec:intro}

\section{BACKGROUND}
We consider the RL setting where the environment is modeled as an infinite-horizon discrete-time Markov Decision Process (MDP). The MDP is characterized by the tuple ($\mathcal{S}$, $\mathcal{A}$, $\mathcal{R}$, $\mathcal{T}$, $\gamma$, $p_0$), where $\mathcal{S}$ and $\mathcal{A}$ are the continuous state- and action-space, respectively, $\gamma\in[0,1)$ is the discount factor, and $p_0$ is the initial state distribution. Given an action $a_t \in \mathcal{A}$, the next state is sampled from the transition dynamics distribution, $s_{t+1} \sim \mathcal{T}(s_{t+1}|s_t,a_t)$, and the agent receives a scalar reward $r(s_t,a_t)$ determined by the reward function $\mathcal{R}$. A policy $\pi_\theta(a_t|s_t)$ defines the state-conditioned distribution over actions. The RL objective is to learn the policy parameters ($\theta$) to maximize the expected discounted sum of rewards, $\eta(\pi_{\theta}) = \mathbb{E}_{p_0, \mathcal{T}, \pi} \big[ \sum_{t=0}^{\infty} \gamma^t r(s_t,a_t) \big]$.

Policy-gradient algorithms~\citep{sutton2000policy} are widely used to estimate the gradient of the RL objective. Proximal policy optimization (PPO,~\citet{schulman2017proximal}) is a model-free policy-gradient algorithm that serves as an efficient approximation to trust-region methods~\citep{schulman2015trust}. In each iteration of PPO, the rollout policy ($\pi_{\theta_{\text{old}}}$) is used to collect sample trajectories $\tau$ and the following surrogate loss is minimized over multiple epochs:
\begin{equation*}
    L^{\theta}_{\text{PPO}} = -\mathbb{E}_{\tau} \Big[ \text{min}\left(r_t\left(\theta\right)\hat{A}_t, \text{clip}\left(r_t\left(\theta\right), 1 - \epsilon, 1 + \epsilon\right)\hat{A}_t\right) \Big]
\end{equation*}
where $r_t(\theta) = \frac{\pi_\theta(a_t|s_t)}{\pi_{\theta_{\text{old}}}(a_t|s_t)}$ is the ratio of the action probabilities under the current policy and rollout policy, and $\hat{A}_t$ is the estimated advantage. Variance in the policy-gradient estimates is reduced by employing the state-value function as a control variate~\citep{mnih2016asynchronous}. This is usually modeled as a neural network $V_\psi$ and updated using temporal difference learning:
\begin{equation*}
    L^{\psi}_{\text{PPO}} = -\mathbb{E}_{\tau} 
\Big[ \left(V_{\psi}\left(s_t\right) - V_t^{\text{targ}}\right)^2 \Big]
\end{equation*}
where $V_t^{\text{targ}}$ is the bootstrapped target value obtained with TD($\lambda$). To further reduce variance, Generalized Advantage Estimation (GAE,~\citet{schulman2015high}) is used when estimating advantage. The overall PPO minimization objective then is:
\begin{equation}\label{eq:ppo_loss}
    L_{\text{PPO}} (\theta, \psi) = L^{\theta}_{\text{PPO}} + L^{\psi}_{\text{PPO}}
\end{equation}
Although we use the PPO objective for our experiments, our method can be readily combined with any on-policy or off-policy actor-critic RL algorithm. 

\label{sec:backgr}

\section{METHOD}

In this section, we outline our method for distilling knowledge from a pre-trained teacher policy to a student policy, in the hope that such knowledge sharing improves the sample-efficiency of the student learning process. Our problem setting is as follows. We assume that the teacher and the student policies operate in two different MDPs. All the MDP properties ($\mathcal{S}$, $\mathcal{A}$, $\mathcal{R}$, $\mathcal{T}$, $\gamma$, $p_0$) could be different, provided that some high-level structural commonality exists between the MDPs, such as the example of transfer from a quadruped robot to hexapod robot introduced in Section~\ref{sec:intro}. Henceforth, for notational convenience, we refer to the MDP of the teacher as the {\em source} MDP, and that of the student as the {\em target} MDP. We assume the availability of a teacher policy network pre-trained in the source MDP. Crucially though, we do not assume access to the source MDP for any further exploration, or for obtaining demonstration trajectories that could be used for training in the target MDP using cross-domain imitation-learning techniques. We instead focus on extracting representations from the teacher policy network which are useful for learning in the target MDP.

In this work, we address knowledge transfer when $\mathcal{S}_{\text{src}} \neq \mathcal{S}_{\text{targ}}$, where $\mathcal{S}_{\text{src}}$ and $\mathcal{S}_{\text{targ}}$ denote the state-space of the source and target MDPs, respectively. To handle the mismatch, we introduce a learned embedding-space parameterized by an encoder function $\phi(\cdot)$, and defined as $\mathcal{S}_{\text{emb}} \coloneqq \{\phi(s) \mid s \in \mathcal{S}_{\text{targ}}\}$. Data points from this embedding space are used to extract useful information from the teacher policy network. Therefore, we further enforce that the dimension of the embedding space matches the dimension of the state-space in the source MDP, i.e.,  $|\mathcal{S}_{\text{emb}}| = |\mathcal{S}_{\text{src}}|$. Note that this does not necessitate that any embedding vector $s \in \mathcal{S}_{\text{emb}}$ be a feasible input state in the source MDP. To learn the encoder function $\phi(\cdot)$, we consider the following two desiderata. Firstly, the embeddings must be learned to facilitate our objective of maximizing the cumulative discount rewards in the target MDP. In subsection~\ref{subsec:task_aligned_space}, we show how to achieve this by utilizing the policy gradient to update embedding parameters. Secondly, we wish for a high correlation between the input states of the target MDP and the embedding vectors produced from them. The embeddings are used to deterministically derive representations from the teacher network, and hence a high correlation helps to obtain the most appropriate teacher guidance for each of the states encountered by the target policy. To this end, we propose a mutual information maximization objective; this is detailed in subsection~\ref{subsec:mi_reg}.


\subsection{TASK-ALIGNED EMBEDDING SPACE}\label{subsec:task_aligned_space}
%
%
This section describes our approach for training the encoder parameters ($\phi$) such that the generated embeddings are aligned with the RL objective. We begin by detailing the architecture that we use for transfer of knowledge from a teacher, pre-trained in source MDP, to a student policy in the target MDP with different state- and action-space. Inspired by the concept of knowledge-flow used in~\citep{liu2019knowledge}, we employ lateral connections between the student and teacher networks, which augment the representations in the layers of the student with useful representations from the layers of the teacher. A crucial benefit of this approach is that since information sharing happens through the hidden layers, the output (action) space of the source and target MDPs can be disparate, as is the scenario in our experiments. It is also quite straightforward to include multiple teachers in this architecture to distill diverse knowledge into a student; we leave this to future work.

We draw out knowledge from both the teacher policy and state-value networks. We denote the teacher policy and value network with $\pi_{\theta'}$  and $V_{\psi'}$, respectively, where the parameters ($\theta', \psi'$) are held fixed throughout the training. Analogously, ($\theta, \psi$) are the trainable parameters for the student policy and value networks. Let $N_\pi$ denote the number of hidden layers in the teacher (and student) policy network, and $N_V$ be the number of hidden layers in the teacher (and student) value network. In general, the teacher and student networks could have a different number of layers, but we assume them to be the same for ease of exposition. 

In the target MDP, the student policy observes a state $s_{\text{targ}} \in \mathcal{S}_{\text{targ}}$, which is fed to the encoder to produce the embedding $\phi(s_{\text{targ}}) \in \mathcal{S}_{\text{emb}}$. Since $|\mathcal{S}_{\text{emb}}| = |\mathcal{S}_{\text{src}}|$, this embedding can be readily passed through the teacher networks to extract $\{z^j_{\theta'}, 1 \leq j \leq N_{\pi}\}$, representing the pre-activation outputs of the $N_{\pi}$ hidden layers of the teacher policy network, and $\{z^j_{\psi'}, 1 \leq j \leq N_{V}\}$, representing the pre-activation outputs of the $N_{V}$ hidden layers of the teacher value function network. To obtain the pre-activation representations in the student networks, we feed in the state $s_{\text{targ}}$ and perform a weighted linear combination of the appropriate outputs with the corresponding pre-activations from the teacher networks. Concretely, to obtain the hidden layer outputs $h^j_{\pi_{\theta}}$ and $h_{V_{\psi}}^j$ at layer $j$ in the student networks, we have the following:
\begin{equation}\label{eq:rep_sharing}
\begin{aligned}    
  h^j_{\pi_{\theta}} = \sigma\left(p^j_{\theta}z^j_{\theta} + (1-p^j_{\theta})z^j_{\theta^{'}}\right) \\
  h_{V_{\psi}}^j = \sigma\left(p^j_{\psi}z^j_{\psi} + (1-p^j_{\psi})z^j_{\psi^{'}}\right)
 \end{aligned}  
\end{equation}
where $\sigma$ is the activation function, and $p^j_{\theta}, p^j_{\psi} \in [0,1]$ are layer-specific learnable parameters denoting the mixing weights. In the target MDP, the student network is optimized for the RL objective $L_{\text{PPO}} (\theta, \psi)$, mentioned in Equation~\ref{eq:ppo_loss}. The outputs of the student policy and value networks, and hence $L_{\text{PPO}}$, depend on the encoder parameters ($\phi$) through the representation sharing (Equation~\ref{eq:rep_sharing}) enabled by the lateral connections stemming from the pre-trained teacher network. Therefore, an intuitive objective for shaping the embeddings such that they become task-aligned is to optimize them using the original RL loss gradient: $\phi \leftarrow \phi - \alpha \nabla_{\phi} L_{\text{PPO}} (\theta, \psi, \phi, \theta', \psi')$. Note that $L_{\text{PPO}}(\cdot)$ now also depends on the fixed teacher parameters ($\theta', \psi'$).

The learnable mixing weights $p^j_{\theta}, p^j_{\psi} \in [0,1]$ control the influence of the teacher's representation on the student outputs -- higher the value, lesser the impact. We argue that a low value for these coefficients helps in the early phases of the training process by providing necessary information to kick-start learning. At the end of the training, however, we desire that the student becomes completely independent of the teacher, since this helps in faster test-time deployment of the agent. To encourage this, we introduce additional {\em coupling-loss} terms that drive $p^j_{\theta}, p^j_{\psi}$ towards 1 as the training progresses:
\begin{equation}\label{eq:coupling_loss}
   L_{\text{coupling}} =  -\frac{1}{N_{\pi}}\sum_{j=1}^{N_{\pi}}\log\left(p^j_{\theta}\right) -\frac{1}{N_{V}}\sum_{j=1}^{N_{V}}\log\left(p^j_{\psi}\right)
\end{equation}
Experimentally, we observe that although the student becomes independent in the final stages of training, it is able to achieve the same level of performance that it would if it could still rely on the teacher.
\subsection{ENRICHED EMBEDDINGS WITH MUTUAL INFORMATION MAXIMIZATION}\label{subsec:mi_reg}
As outlined in the previous section, at each timestep of the discrete-time target MDP, the representation distilled from the teacher networks is a fixed function $f$ of the embedding vector generated from the current input state: $f(\theta',\psi', \phi(s_{\text{targ}}))$, where ($\theta',\psi'$) are fixed. It is desirable to have a high degree of correlation between $s_{\text{targ}}$ and $f(\theta',\psi', \phi(s_{\text{targ}}))$ because, intuitively, the teacher representation that is the most useful for the student should be different at different input states. To aid with this, we utilize a surrogate objective that instead maximizes the correlation between $s_{\text{targ}}$ and the embeddings $\phi(s_{\text{targ}})$, defined using the principle of mutual information (MI). If we view $s_{\text{targ}}$ as a stochastic input $\boldsymbol{s}$, the encoder output is then also a random variable $\boldsymbol{e}$, and the mutual information between the two is defined as:
\begin{equation*}
    \mathcal{I}(\boldsymbol{s};\boldsymbol{e}) = \mathcal{H}(\boldsymbol{s}) - \mathcal{H}(\boldsymbol{s} | \boldsymbol{e}) 
\end{equation*}
where $\mathcal{H}$ denotes the differential entropy. Direct maximizing of the MI is intractable due to the unknown conditional densities. However, it is possible to obtain a lower bound to the MI using a variational distribution $q_\omega(\boldsymbol{s} | \boldsymbol{e})$ that approximates the true conditional distribution $p(\boldsymbol{s} | \boldsymbol{e})$ as follows:
\begin{equation*}
\begin{aligned}
     \mathcal{I}(\boldsymbol{s};\boldsymbol{e}) &= \mathcal{H}(\boldsymbol{s}) - \mathcal{H}(\boldsymbol{s} | \boldsymbol{e}) \\
     &= \mathcal{H}(\boldsymbol{s}) + \mathbb{E}_{\boldsymbol{s}, \boldsymbol{e}} [\log p(\boldsymbol{s} | \boldsymbol{e})] \\ 
    &= \mathcal{H}(\boldsymbol{s}) + \mathbb{E}_{\boldsymbol{s}, \boldsymbol{e}} [\log q_\omega(\boldsymbol{s} | \boldsymbol{e})] \\
    &\quad \quad + \mathbb{E}_{\boldsymbol{e}}\big[ D_{\text{KL}} (p(\boldsymbol{s} | \boldsymbol{e})||q_\omega(\boldsymbol{s} | \boldsymbol{e}))\big] \\ 
    &\ge \mathcal{H}(\boldsymbol{s}) + \mathbb{E}_{\boldsymbol{s}, \boldsymbol{e}} [\log q_\omega(\boldsymbol{s} | \boldsymbol{e})]
\end{aligned}
\end{equation*}
where the last inequality is due to the non-negativity of the KL divergence. This is known as the variational information  maximization algorithm~\citep{agakov2004variational}. Re-writing in terms of target-MDP states and the encoder parameters, the surrogate objective jointly optimizes over the variational and encoder parameters:
\begin{equation*}
    \max_{\omega,\phi} \mathbb{E}_{s_{\text{targ}}} [\log q_\omega(s_{\text{targ}}|\phi(s_{\text{targ}}))]
\end{equation*}
where $\mathcal{H}(\boldsymbol{s})$ is omitted since it is a constant {\em w.r.t} the concerned parameters. In terms of the loss function to minimize, we can succinctly write:
\begin{equation}\label{eq:mi_loss}
    L_{\text{MI}} (\phi, \omega) = - \mathbb{E}_{s\sim\rho_{\pi_\theta}}[\log q_\omega(s|\phi(s))]
\end{equation}
where $\rho_{\pi_\theta}$ is the state-visitation distribution of the student policy in the target MDP. In our experiments, we use a multivariate Gaussian distribution (with a learned diagonal covariance matrix) to model the variational distribution $q_\omega$. Although this simple model yields good performance, more expressive model classes, such as mixture density networks and flow-based models~\citep{rezende2015variational} could be readily incorporated as well, to learn complex and multi-modal distributions.
\subsection{OVERALL ALGORITHM}
  \begin{figure*}[t]
  \centering
 \includegraphics[scale=0.58]{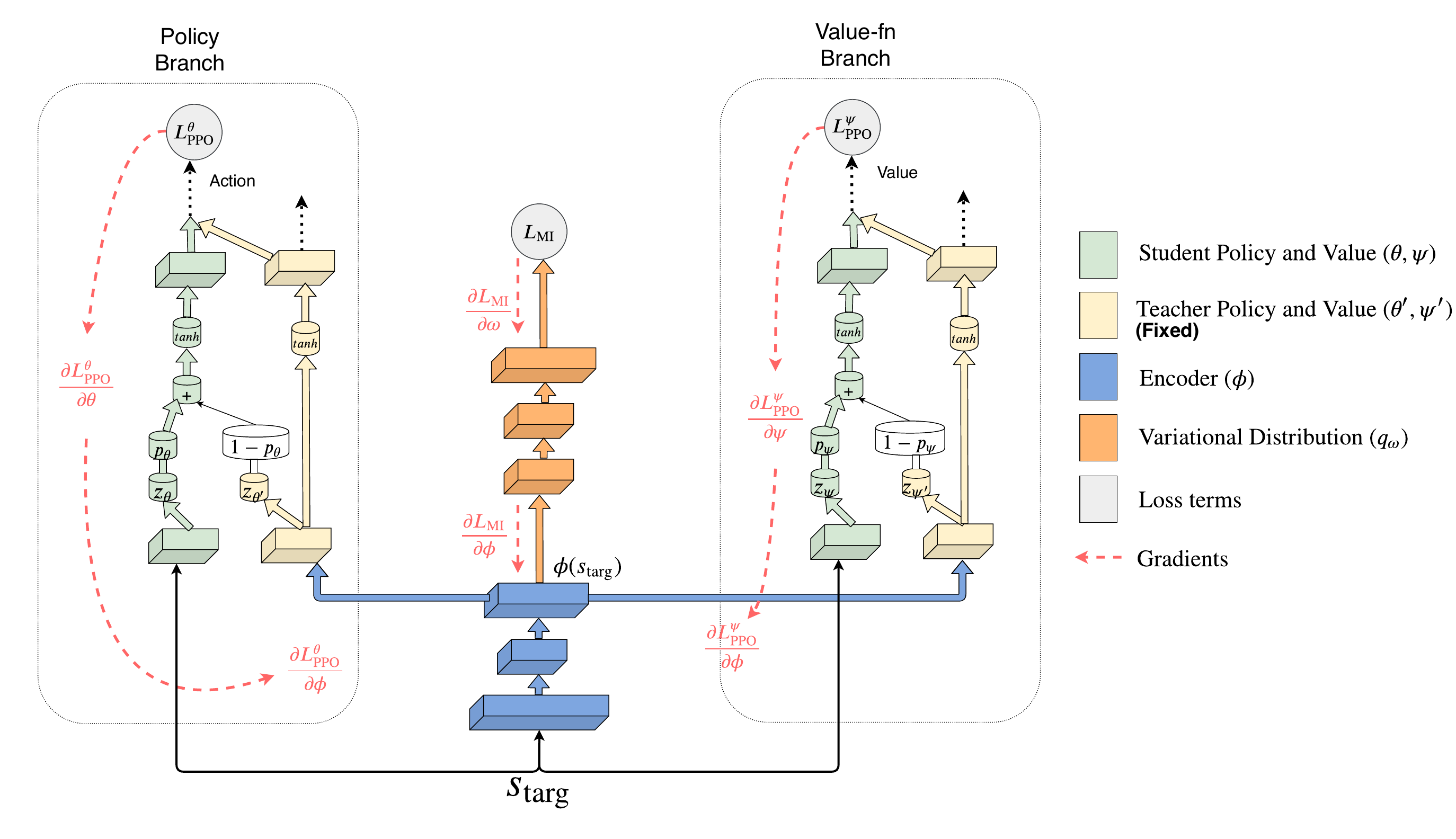}
 \caption{Schematic diagram of our complete architecture (best viewed in color). The encoder parameters $\phi$ (blue) receive gradients from three sources: the policy-gradient loss $L^\theta_{\text{PPO}}$, the value function loss $L^\psi_{\text{PPO}}$, and the mutual information loss $L_{\text{MI}}$. The teacher networks (yellow) remain fixed throughout training and do not receive any gradients. In the student networks (green), the pre-activation representations are linearly combined (using learnt mixing weights) with the corresponding representations from the teacher (Equation~\ref{eq:rep_sharing}). Note that this knowledge-flow occurs at all layers, although we show it only once for clarity of exposition.}
 \label{fig:overall_arch}
 \end{figure*}
Figure~\ref{fig:overall_arch} shows the schematic diagram of our complete architecture and gradient flows, along with a description of the implemented neural networks. We refer to our algorithm as MIKT, for {\em Mutual Information based Knowledge Transfer}. Algorithm~\ref{algo:complete} outlines the main steps of the training procedure. 
%
%
\RestyleAlgo{ruled}
\LinesNumbered
\begin{algorithm}[h!]
\SetNoFillComment
\let\oldnl\nl
\newcommand{\nonl}{\renewcommand{\nl}{\let\nl\oldnl}}
\SetKwInOut{Input}{Input}\SetKwInOut{Output}{Output}
\Input{$\theta', \psi'$ fixed teacher policy and value networks}
\nonl{$\theta, \psi$: student policy and value networks}\\
\nonl{$\{p\}$: set of coupling parameters for policy and value networks}\\
\nonl{$\phi$: encoder parameters}\\
\nonl{$\omega$: variational distribution parameters}\\

 \BlankLine
 \For{each iteration}{
    Run $\pi_\theta$ in target MDP and collect few trajectories $\tau$ \\
    \BlankLine
    \For{each minibatch $m \in \tau$}{
    Update $\theta, \psi$ with $\nabla_{\theta, \psi} L_{\text{PPO}} (\theta, \psi, \phi, \theta', \psi')$ \\ 
    Update $\phi$ with $\nabla_{\phi} \big[ L_{\text{MI}} (\phi, \omega) + L_{\text{PPO}} (\theta, \psi, \phi, \theta', \psi')\big]$ \\
    Update $\omega$ with $\nabla_{\omega} L_{\text{MI}} (\phi, \omega)$ \\ 
    Update $\{p\}$ using $[ L_{\text{coupling}} + L_{\text{PPO}}]$
    }
 }
 \caption{Mutual Information based Knowledge Transfer (MIKT)}
 \label{algo:complete}
\end{algorithm}
In each iteration, we run the policy in the target MDP and collect a batch of trajectories. This experience is then used to compute the RL loss (Equation~\ref{eq:ppo_loss}) and the mutual information loss (Equation~\ref{eq:mi_loss}), enabling the calculation of gradients for the different parameters (Lines 3--6). Using both the losses to update the encoder ($\phi$) helps us to satisfy the desiderata on the embeddings -- that they should be task-aligned and correlated with the states in the target MDP. The coupling parameters $\{p^j\}$, used for the weighted combination of the representations in the teacher and student networks, are updated with the coupling-loss (Equation~\ref{eq:coupling_loss}) along with the RL loss. In each iteration of the algorithm, the PPO update ensures that the state-action visitation distribution of the policy $\pi_\theta$ is modified by only a small amount. This is because of the clipping on the importance-sampling ratio (Section~\ref{sec:backgr}) when obtaining the PPO gradient. In addition to this, we experimentally found that enforcing an explicit KL-regularization on the policy further stabilizes learning. Let $\pi_\theta, \pi_{\theta_{\text{old}}}$ denote the current and the rollout policy, respectively. The loss is then formalized as:
\begin{equation*}
    L_{\text{KL}} (\theta, \theta_{\text{old}}) =  \mathbb{E}_{s \sim \rho_{\pi_{\theta_{\text{old}}}}} \big[
    \mathcal{D}_{KL}(\pi_{\theta}(\cdot | s) ||  \pi_{\theta_{\text{old}}}(\cdot | s)) \big]
\end{equation*}
\begin{figure*}[t]
\centering
\captionsetup[subfigure]{justification=centering}
    \begin{subfigure}{0.16\textwidth}
    \centering
        \includegraphics[scale=0.13]{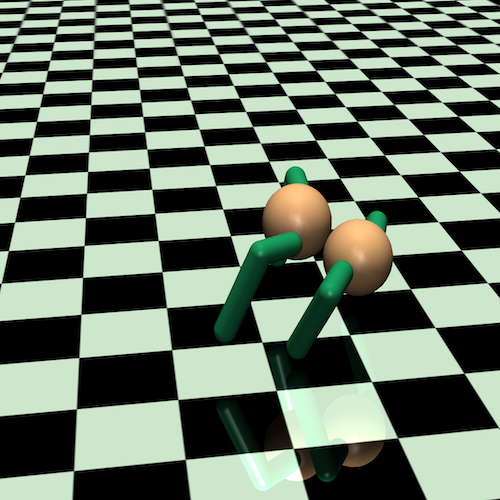}
        \caption{\small{CentipedeFour}}
        \label{}
    \end{subfigure}\hfill%
    \begin{subfigure}{0.16\textwidth}
    \centering
        \includegraphics[scale=0.13]{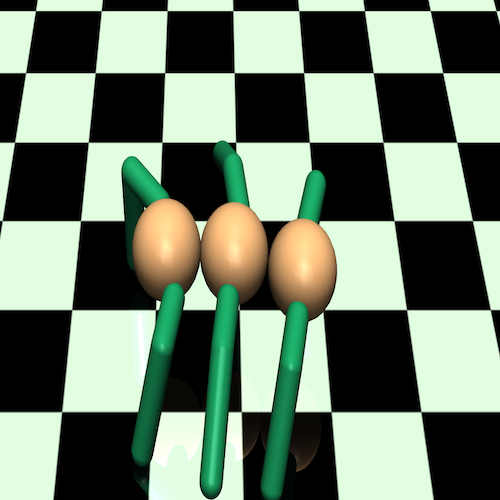}
        \caption{\small{CentipedeSix}}
        \label{}
    \end{subfigure}\hfill%
    \begin{subfigure}{0.16\textwidth}
    \centering
        \includegraphics[scale=0.13]{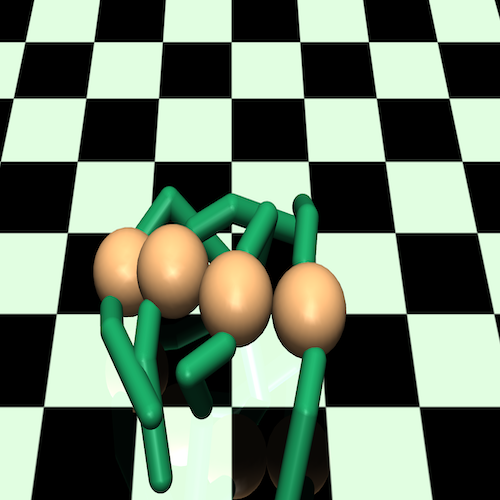}
        \caption{\small{CentipedeEight}}
        \label{}
    \end{subfigure}\hfill%
    \begin{subfigure}{0.16\textwidth}
    \centering
        \includegraphics[scale=0.13]{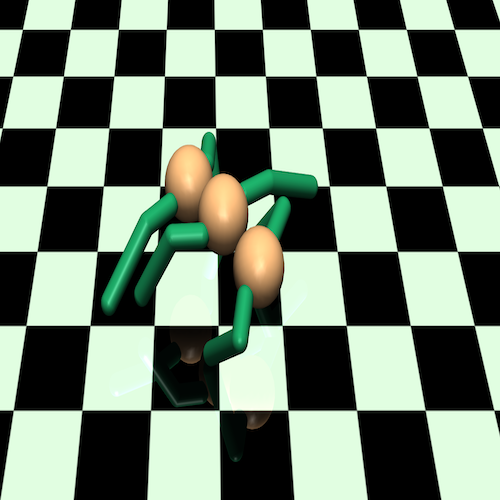}
        \caption{\small{CpCentipedeSix}}
        \label{}
    \end{subfigure}\hfill%
    \begin{subfigure}{0.18\textwidth}
    \centering
        \includegraphics[scale=0.13]{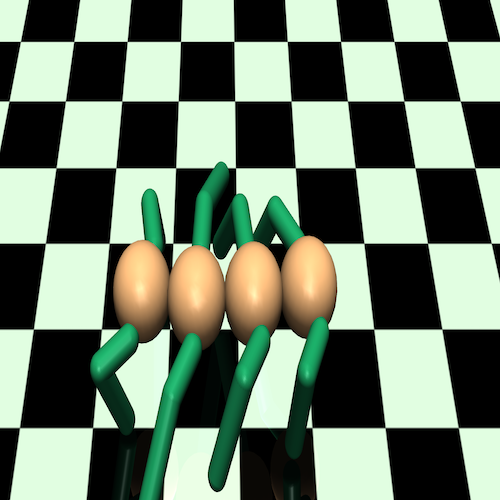}
        \caption{\small{CpCentipedeEight}}
        \label{}
    \end{subfigure}\hfill%
    \begin{subfigure}{0.16\textwidth}
    \centering
        \includegraphics[scale=0.13]{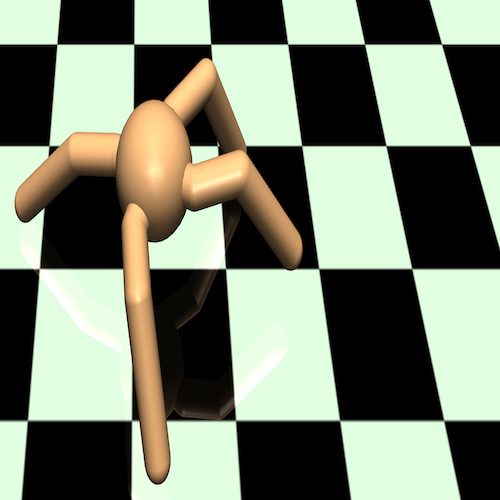}
        \caption{\small{Ant}}
        \label{}
    \end{subfigure}\hfill
    \caption{\small {Our MuJoCo locomotion environments. The centipede agents are configured using the details in~\citep{wang2018nervenet}, while Ant-v2 is a popular OpenAI Gym task.}}
    \label{fig:environments}
    
\end{figure*}

\section{RELATED WORK}
The concepts of knowledge transfer and information sharing between deep neural networks have been extensively researched for a wide variety of tasks in machine learning. In the context of reinforcement learning, the popular paradigms for knowledge transfer include imitation-learning, meta-RL, and policy distillation; each of these being applicable under different settings and assumptions. Imitation learning algorithms~\citep{ng2000algorithms, ziebart2008maximum} utilize teacher demonstrations to extract useful information (such as the teacher reward function in inverse-RL methods) and use that to accelerate student learning. In meta-RL approaches~\citep{duan2016rl, finn2017model}, we are generally provided with a distribution of tasks that share some structural similarity, and the objective is to discover this generalizable knowledge for accelerating the process of learning on a new task. Our work is most closely related to policy distillation methods~\citep{rusu2015policy, parisotto2015actor, czarnecki2019distilling}, where pre-trained teacher networks are available and can expedite learning in dissimilar (but related) student tasks. 

Prior work has considered teachers in various capacities.~\citet{rusu2016progressive} and~\citet{liu2019knowledge} utilize learned cross-connections between intermediate layers of teacher networks---that have been pre-trained on various source tasks---and a student network to effectively transfer knowledge and enable more efficient learning on a target task.~\citet{ahn2019variational} use an objective based on the mutual information between the corresponding layers of teacher and student networks, and show gains in image classification tasks. In~\citet{hinton2015distilling}, information from a large model (teacher) is compressed into a smaller model (student) using a distillation process that uses the temperature-regulated softmax outputs from the teacher as targets to train the student.~\citet{schmitt2018kickstarting} propose a large-scale population-based training pipeline that allows a student policy to leverage multiple teachers specialized in different tasks. All these aforementioned methods work in the setting where the teacher and student share the input state (observation) space. Different from these, our approach handles the mismatch in the state-space by training an embedding space which is utilized for efficient knowledge transfer.~\citet{rozantsev2018beyond} employ layer-wise weight regularization and evaluate on (un-)supervised tasks where the input distributions for source and target domains have semantic similarity and are static. For RL tasks, the input distributions change dynamically as the student policy updates; it is unclear if enforcing similarity between the networks for all inputs by coupling the weights is ideal.~\citet{gamrian2018transfer} use GANs to learn a mapping from target states to source states. In addition to requiring that the source and the target domains have the same action-space, their method also relies on the exploratory samples collected in the source MDP for training the GAN. In contrast, we handle the action-space mismatch and do not assume access to the source MDP for exploration.

Our work also has connections to policy distillation methods that use {\em implicit} teachers, rather than external pre-trained models. In~\citet{czarnecki2018mix}, the authors recommend a curriculum over agents, rather than the usual curriculum over tasks. Such a curriculum trains simple agents first, the knowledge of which is then distilled into more complex agents over time.~\citet{akkaya2019solving} iterate on policy architectures by utilizing behavior-cloning with DAgger; the new architecture (student) is trained using the old architecture (teacher). Distillation has been used in multi-task RL~\citep{teh2017distral} to learn a centralized policy that captures generalizable information from policies trained on individual tasks.~\citep{gangwani2017policy} combine ideas from the genetic-algorithms literature and distillation to train offspring policies that inherit the best traits of both the parent policies. Since all these approaches transfer information in the action-space by minimizing the KL-divergence between state-conditional action distributions, they share the limitation that the student can only leverage a teacher with the same output (action) space. Our approach avoids this by using the representations in the different layers of the neural network for knowledge sharing, enabling transfer-learning in many diverse scenarios as shown in our experiments.
\begin{figure*}[t]
\minipage{0.33\textwidth}
\includegraphics[width=\textwidth, height=4cm]{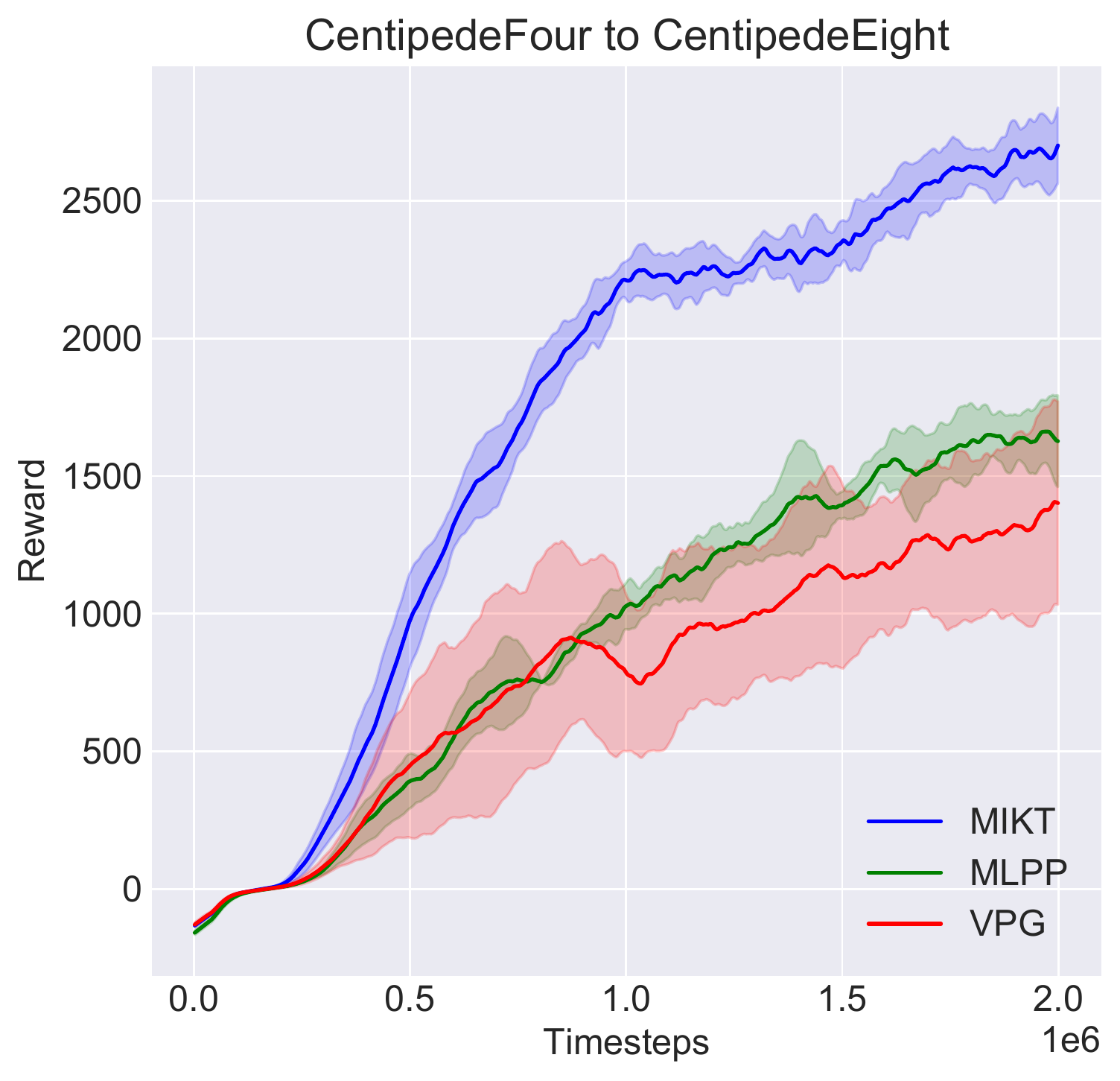} \\ \centering (a)
\endminipage\hfill
\minipage{0.33\textwidth}
\includegraphics[width=\textwidth, height=4cm]{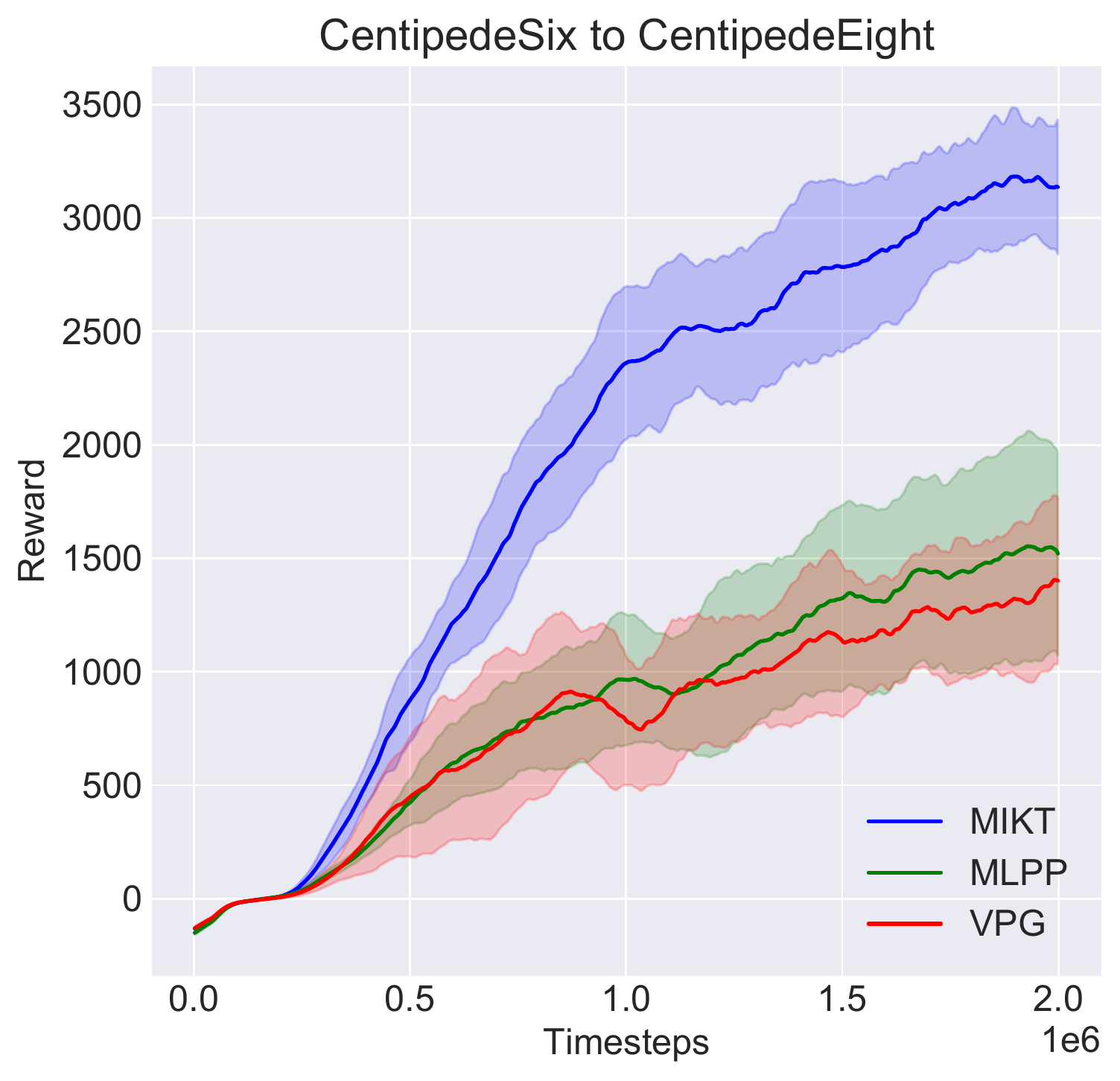} \\ \centering (b)
\endminipage\hfill
\minipage{0.33\textwidth}
\includegraphics[width=\textwidth, height=4cm]{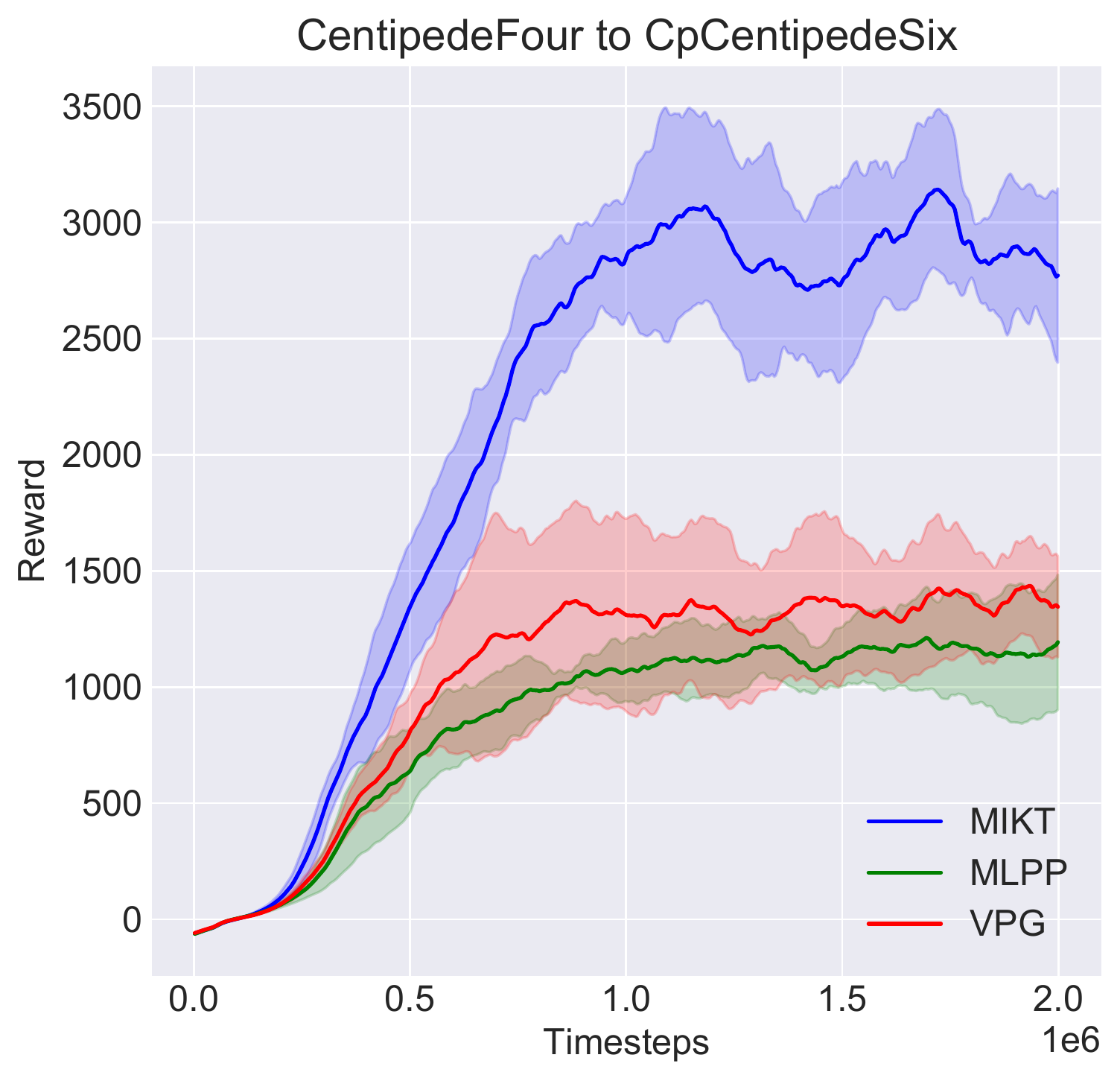} \\ \centering (c)
\endminipage\hfill
\minipage{0.33\textwidth}
\includegraphics[width=\textwidth, height=4cm]{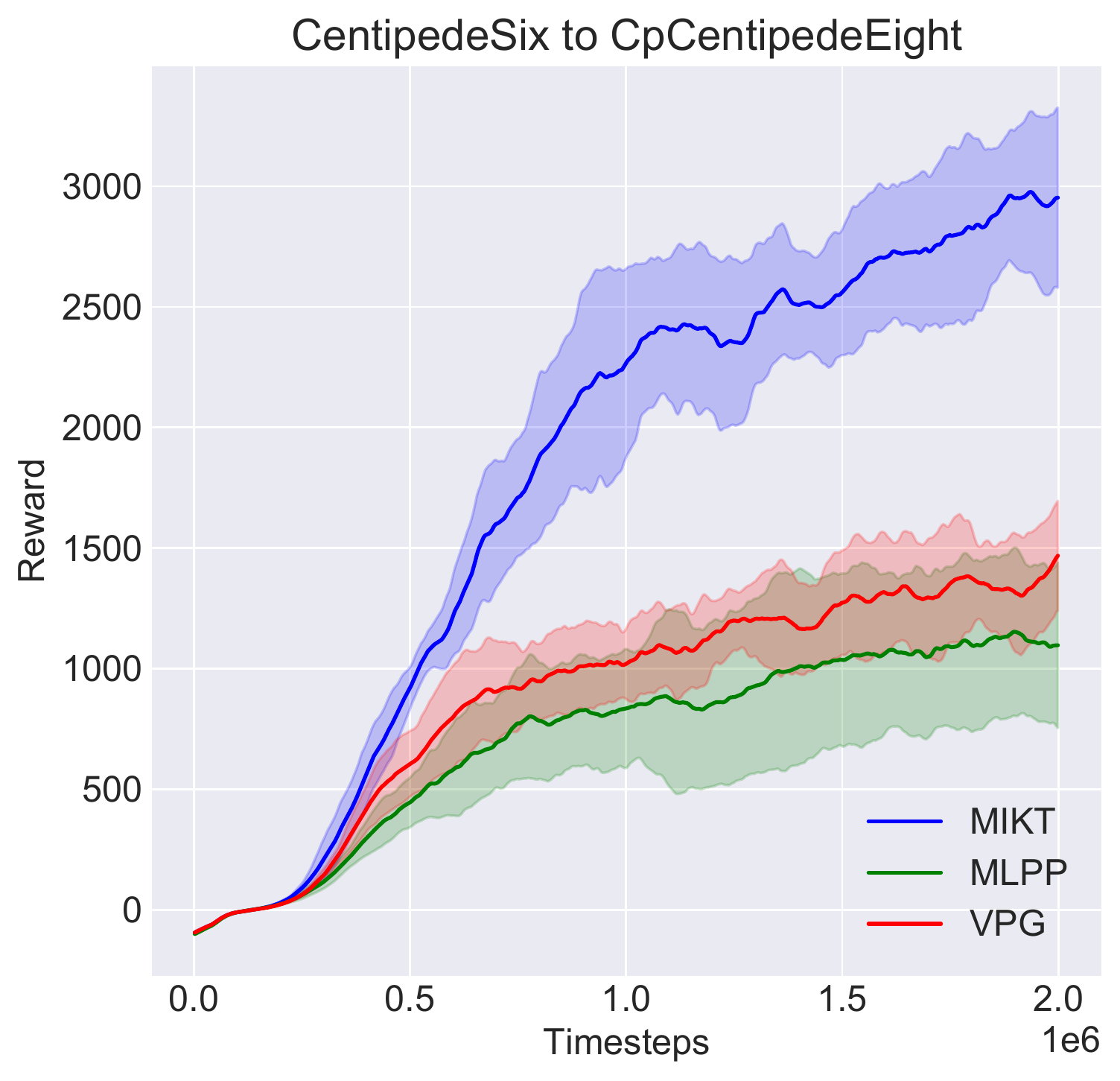} \\ \centering (d)
\endminipage\hfill
\minipage{0.33\textwidth}
\includegraphics[width=\textwidth, height=4cm]{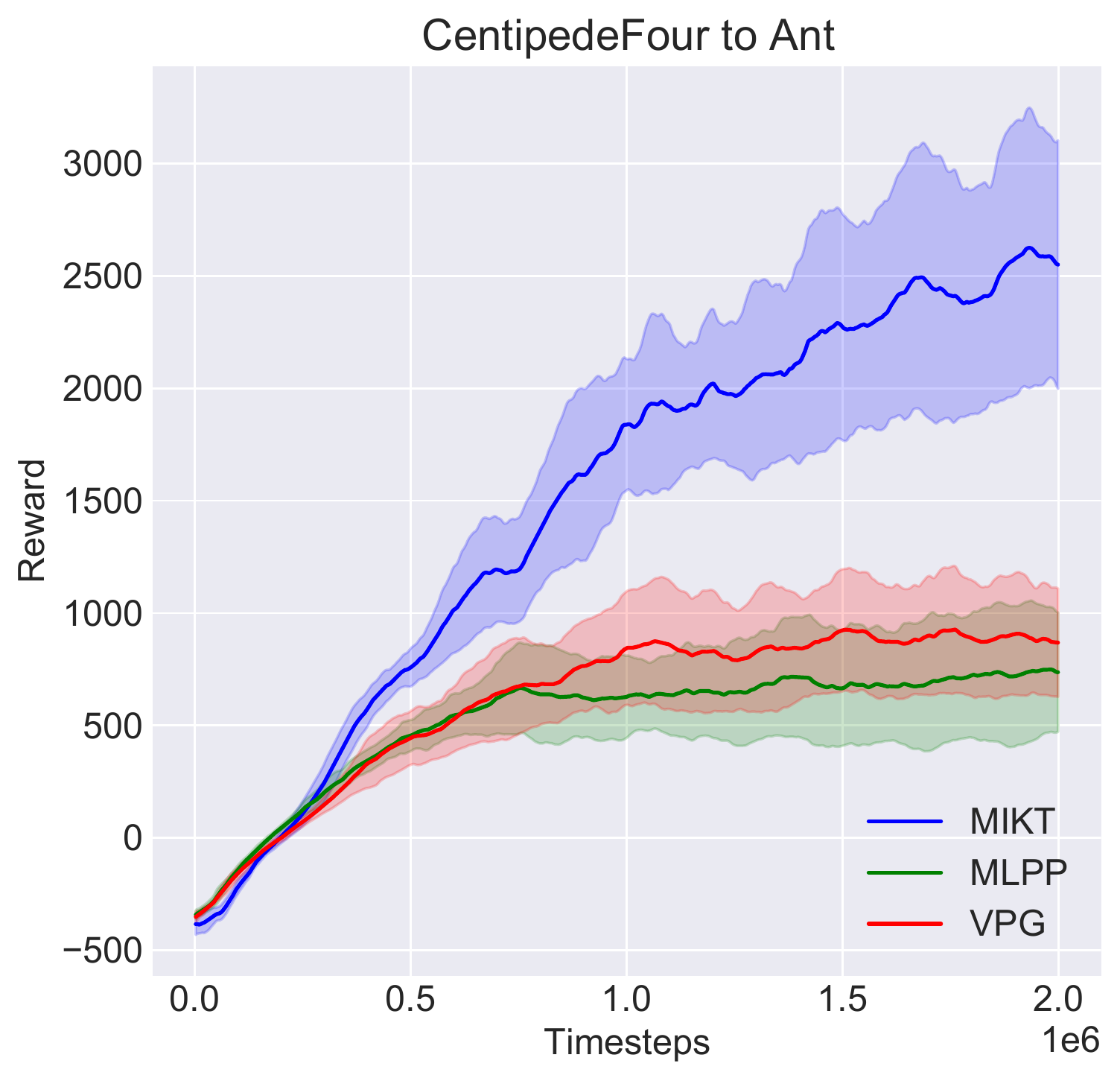} \\ \centering (e)
\endminipage\hfill
\minipage{0.33\textwidth}
\includegraphics[width=\textwidth, height=4cm]{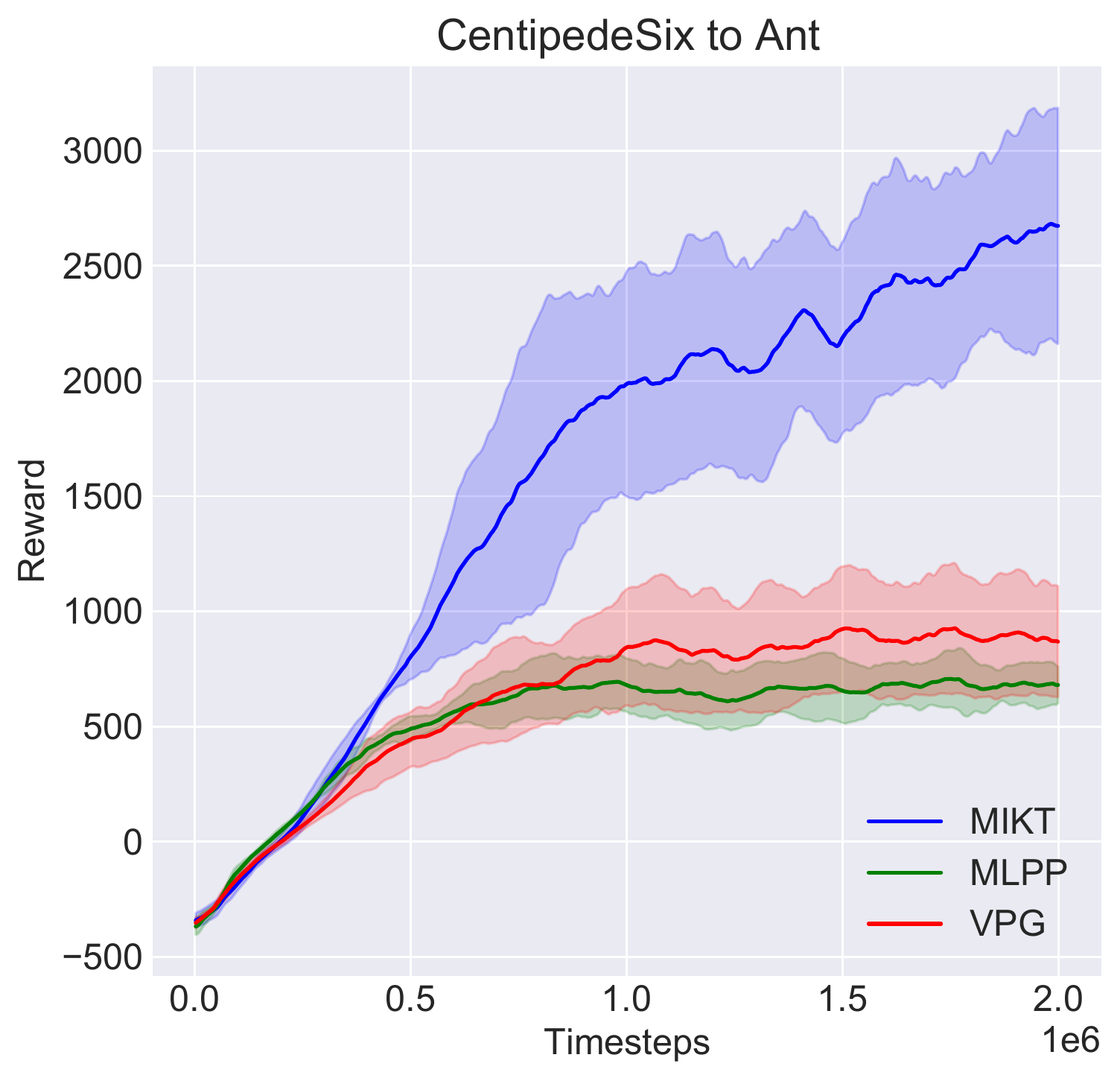} \\ \centering (f)
\endminipage\hfill
\caption{Performance of our transfer learning algorithm (MIKT) and the baselines (VPG, MLPP) on the MuJoCo locomotion tasks. Each plot is titled ``$x$ to $y$", where $x$ is the source (teacher) MDP and $y$ is the target (student) MDP. (a) CentipedeFour to CentipedeEight, (b) CentipedeSix to CentipedeEight, (c) CentipedeFour to CpCentipedeSix, (d) CentipedeSix to CpCentipedeEight, (e) CentipedeFour to Ant, (f) CentipedeSix to Ant.}
\label{fig:overall_perf}
\end{figure*}
%


\section{EXPERIMENTS}
%
%
%
%
\begin{figure*}[t]
\minipage{0.33\textwidth}
\includegraphics[width=\textwidth, height=4cm]{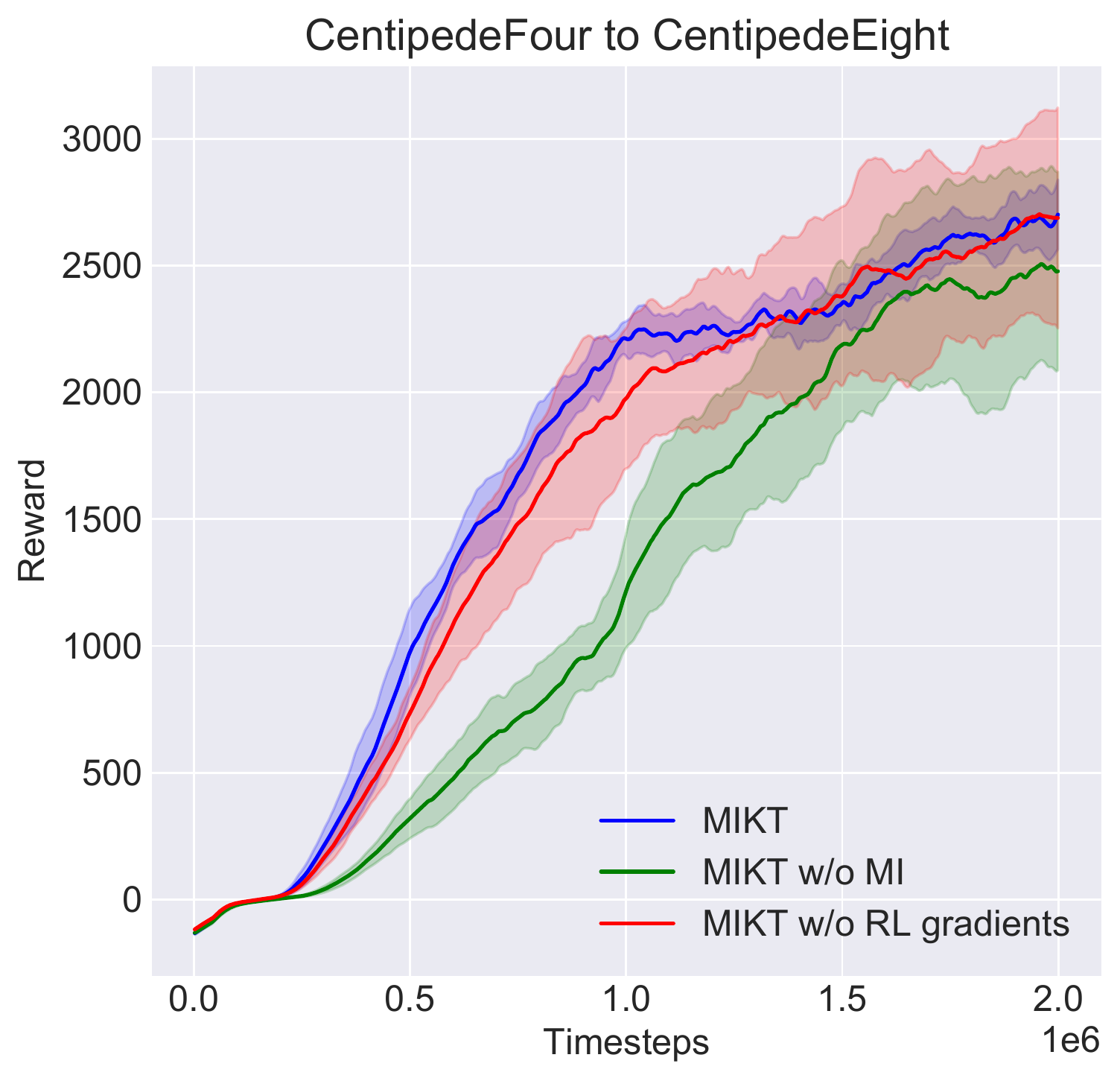} \\ \centering (a)
\endminipage\hfill
\minipage{0.33\textwidth}
\includegraphics[width=\textwidth, height=4cm]{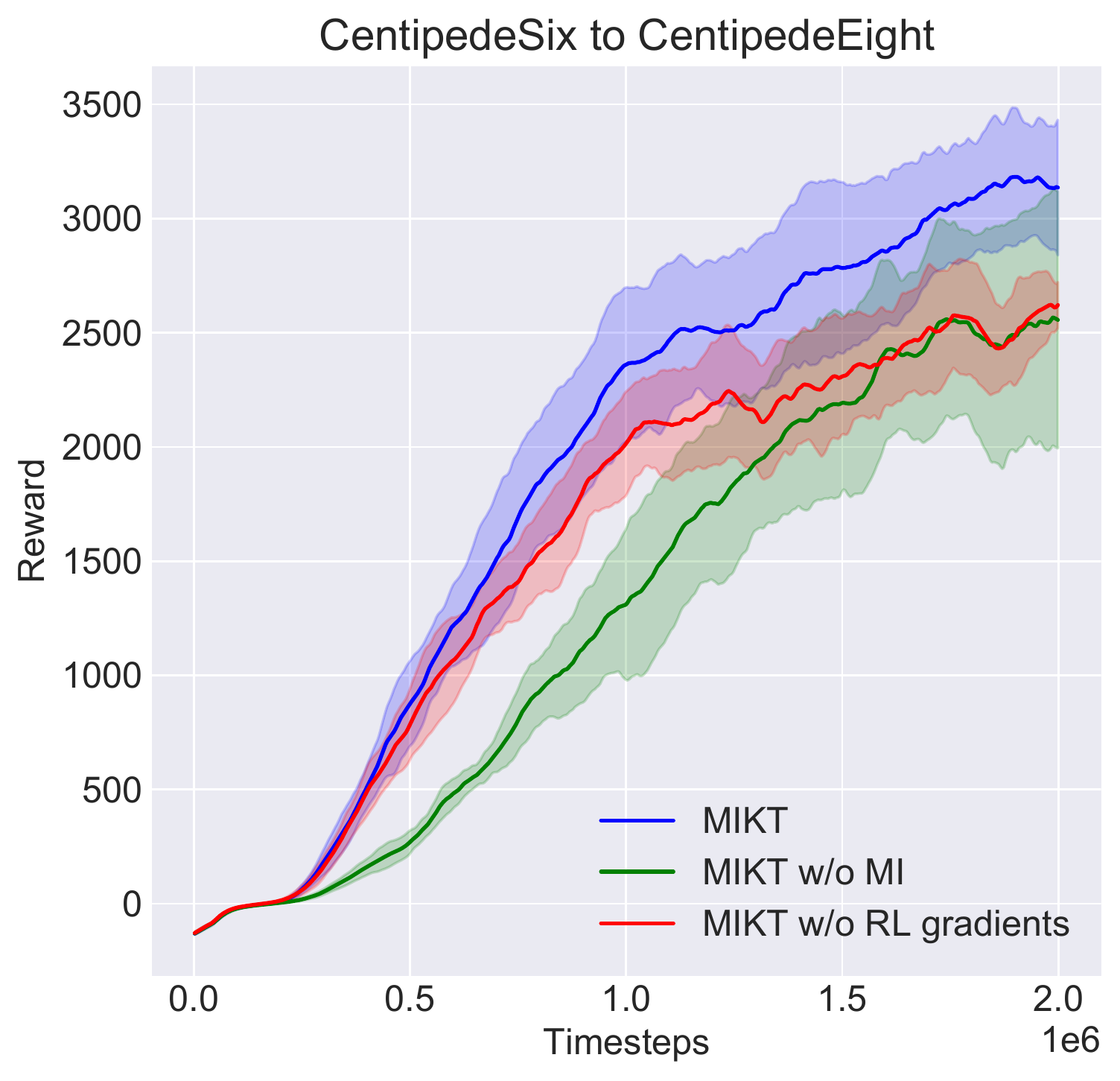} \\ \centering (b)
\endminipage\hfill
\minipage{0.33\textwidth}
\includegraphics[width=\textwidth, height=4cm]{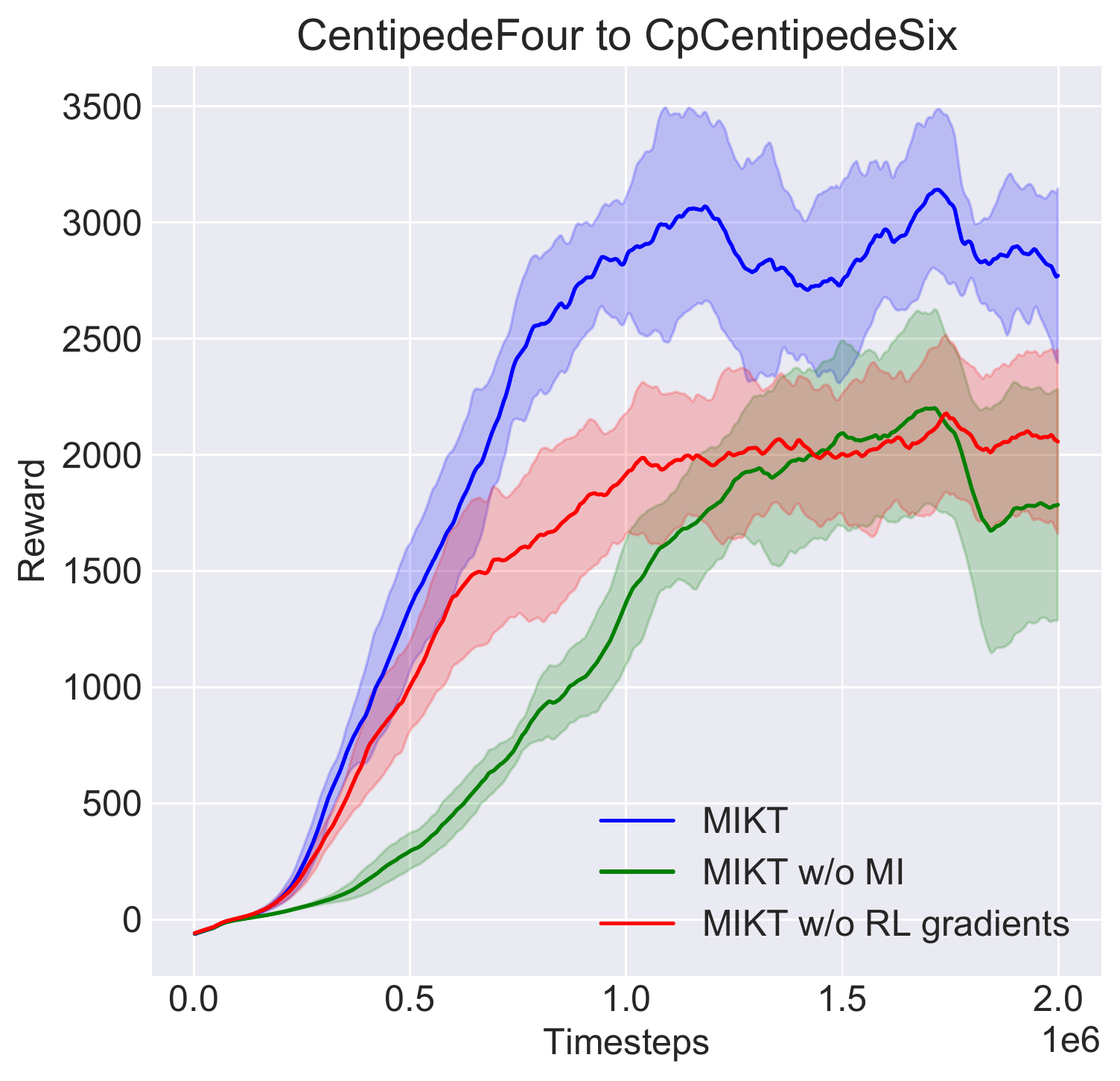} \\ \centering (c)
\endminipage\hfill
\minipage{0.33\textwidth}
\includegraphics[width=\textwidth, height=4cm]{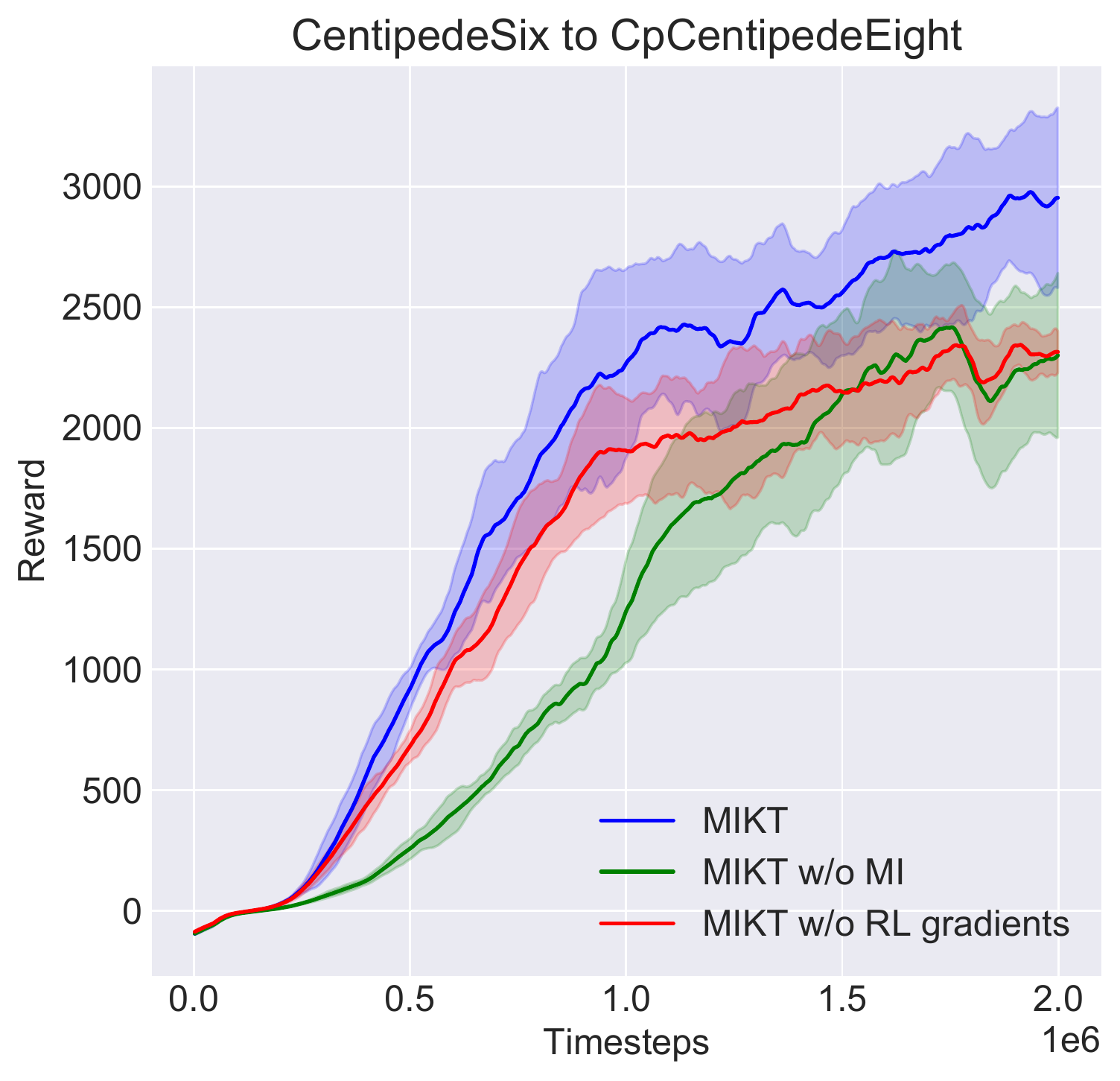} \\ \centering (d)
\endminipage\hfill
\minipage{0.33\textwidth}
\includegraphics[width=\textwidth, height=4cm]{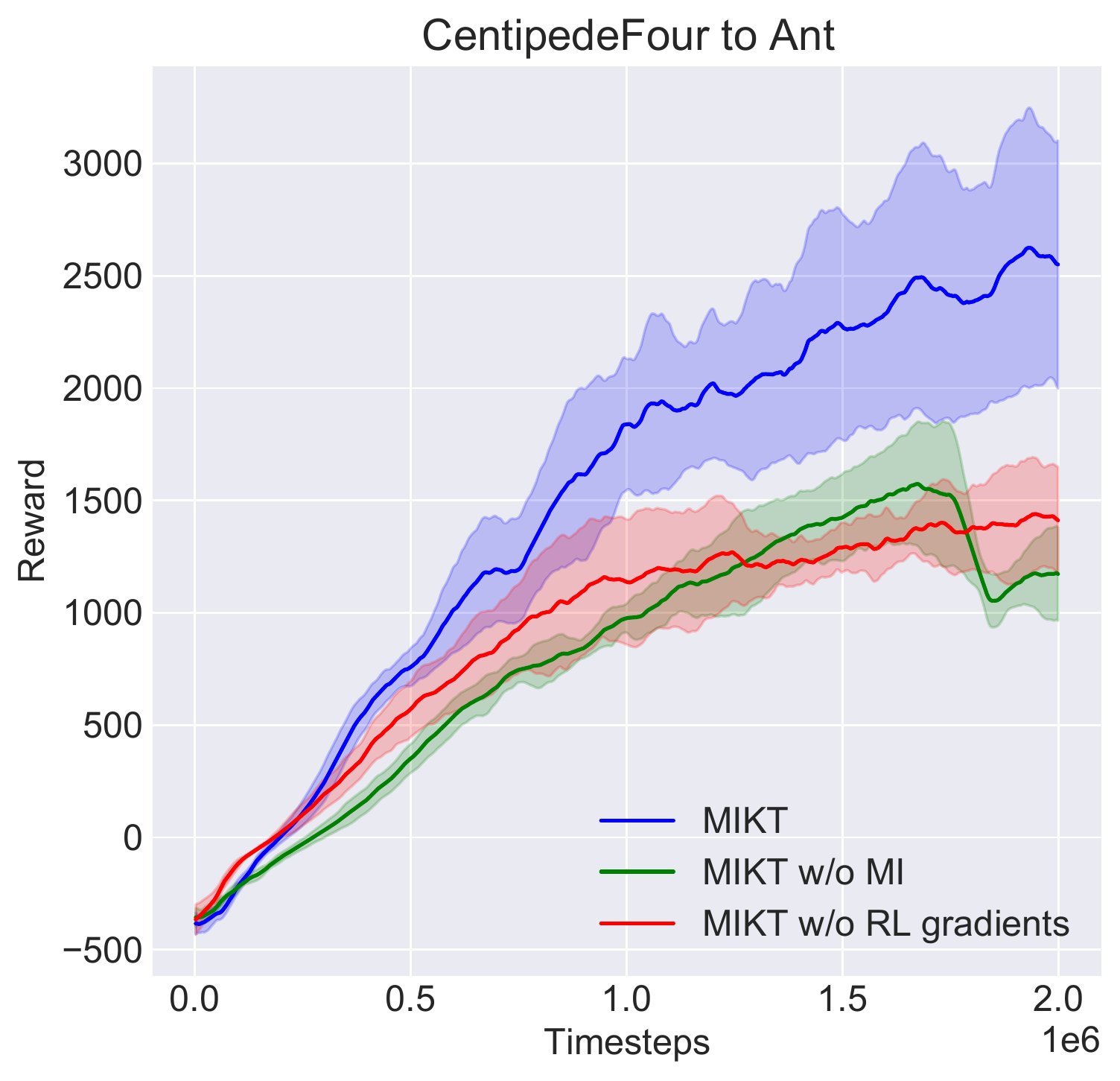} \\ \centering (e)
\endminipage\hfill
\minipage{0.33\textwidth}
\includegraphics[width=\textwidth, height=4cm]{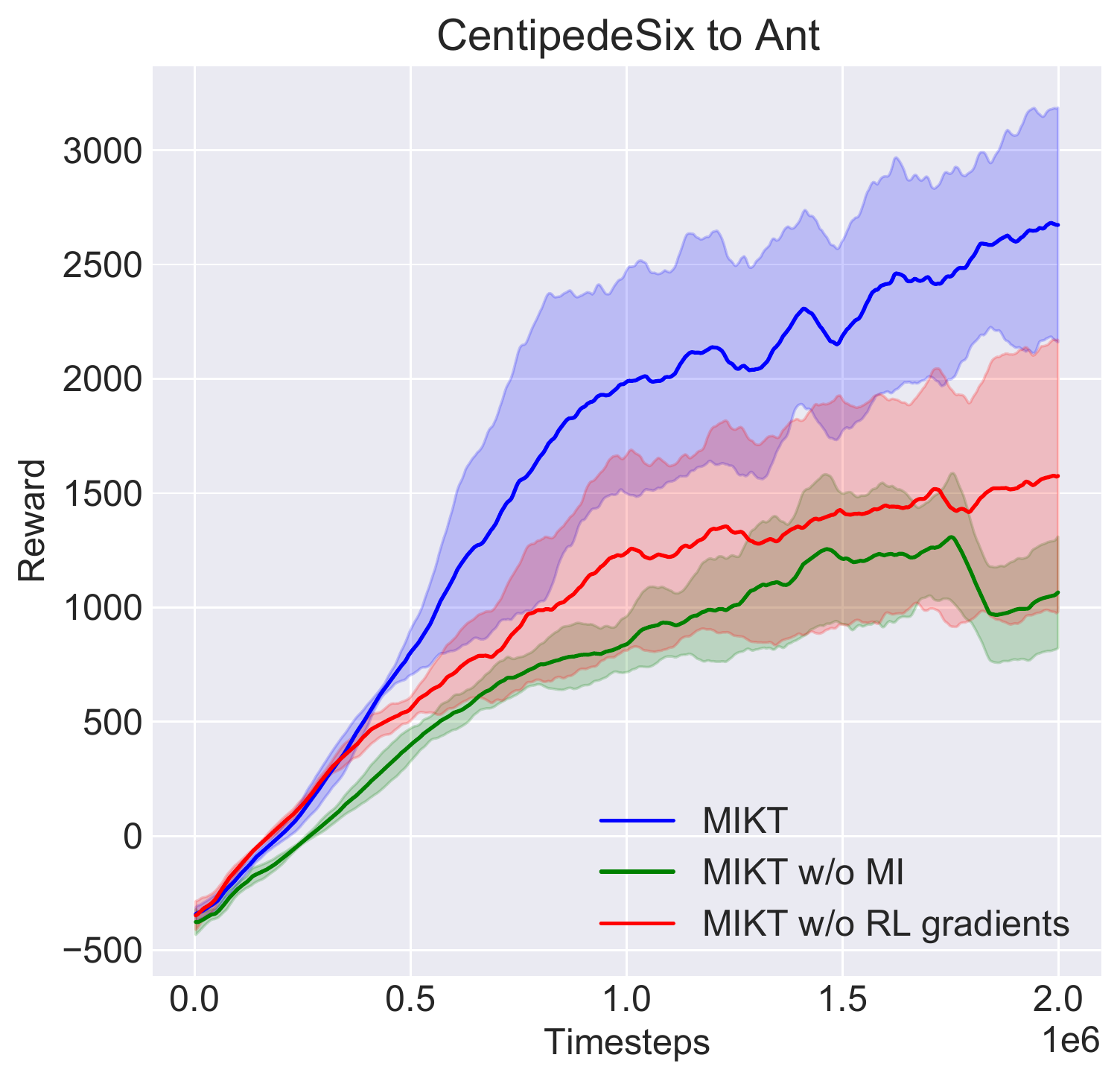} \\ \centering (f)
\endminipage\hfill
\caption{Ablation on the importance of each of $\{L_{\text{PPO}}, L_{\text{MI}}\}$ for training the encoder $\phi$. MIKT (blue) is compared with two variants: {\em MIKT w/o MI} ($L_{\text{MI}}$ not used) and {\em MIKT w/o RL gradients} ($L_{\text{PPO}}$ not used).}
\label{fig:ablations}
\end{figure*}
In this section, we perform experiments to quantify the efficacy of our algorithm, MIKT, for transfer learning in RL, and also do some qualitative analysis. We address the following questions: a) Can we do successful knowledge transfer between a teacher and a student with {\em different state- and action-space}? b) Are both the losses $\{L_{\text{PPO}}, L_{\text{MI}}\}$ important for learning useful embeddings $\phi$? c) How does task-similarity affect the benefits that can be reaped from MIKT?  

\begin{table}[h]
\centering
\caption{\small MuJoCo locomotion environments.}
\resizebox{\linewidth}{!}{%
\begin{tabular}{c|c|c}
                       Environment&State Dimension&Action Dimension\\ \hline \hline
                       
    CentipedeFour & 97 & 10\\
    CentipedeSix & 139 & 16\\
    CentipedeEight & 181 & 22\\
    CpCentipedeSix & 139 & 12\\
    CpCentipedeEight & 181 & 18\\
    Ant & 111 & 8\\   
\end{tabular}}
\label{table:environments}
\end{table}

\textbf{Environments:} We evaluate using locomotion tasks for legged robots, modeled in OpenAI Gym~\citep{gymbrockman} using the MuJoCo physics simulator. Specifically, we use the environments provided by~\citet{wang2018nervenet}, where the agent structure resembles that of a centipede -- it consists of repetitive torso bodies, each having two legs. Figure~\ref{fig:environments} shows an illustration of the different centipede agents. Please see~\citep{wang2018nervenet} for a detailed description of the environment generation process. The agent is rewarded for running fast in a particular direction. Table~\ref{table:environments} includes the state and action dimensions of all the agents. Centipede-$x$ refers to a centipede with $x$ legs; we use $x \in \{4,6,8\}$. We use additional environments where the centipede is crippled (some legs disabled) and denote this by  {\em Cp-}Centipede-$x$. Finally, we include the standard Ant-v2 task from the MuJoCo suite. Note that all robots have separate state and action dimensions. Intuitively though, these locomotion tasks share an inherent structure that could be exploited for transfer learning between the centipedes of various types. We now demonstrate that our algorithm achieves this successfully.
%
%

\textbf{Baselines:} We compare MIKT with two baselines: a) {\em Vanilla Policy Gradient (VPG)}, which learns the task in the target MDP from scratch using only the environmental rewards. Any transfer learning algorithm which effectively leverages the available teacher networks should be able to outperform this baseline that does not receive any prior knowledge it can use. We use the standard PPO~\citep{schulman2017proximal} algorithm for this baseline. b) {\em MLP Pre-trained (MLPP)} In our setting, the teacher and the student networks have dissimilar input and output dimensions (because the MDPs have different state- and action-spaces). A natural transfer learning strategy is to remove the input and output layers from the pre-trained teacher and replace them with new learnable layers that match the dimensions required of the student policy (analogously value) network. The middle stack of the deep neural network is then fine-tuned with the RL loss. Prior work has shown that such a transfer is effective in certain computer vision tasks.

%
\subsection{EXPERIMENTAL RESULTS}
Figure~\ref{fig:overall_perf} plots the learning curves for MIKT and our two baselines in different transfer learning experiments. Each plot is titled ``$x$ to $y$", where $x$ is the source (teacher) MDP and $y$ is the target (student) MDP. We run each experiment with 5 different random seeds and plot the average episodic returns (mean and standard deviation) on the y-axis, against the number of timesteps of environment interaction (2 million total) on the x-axis. VPG does not use utilize the pre-trained teachers. We observe that its performance improves with the training iterations, albeit at a sluggish pace. MLPP uses the middle stack of the pre-trained teacher network as an initialization and trains the input and output layers from scratch. It only performs on par with VPG, potentially due to the non-constructive interaction between the pre-trained and randomly initialized parameters of the student networks. This indicates that the MLPP strategy is not productive for transfer learning across the RL locomotion tasks considered. Finally, we note that our algorithm (MIKT) vastly outperforms the two baselines, both achieving higher returns in earlier stages of training and reaching much higher final performance. This proves that firstly, these tasks do have a structural commonality such that a teacher policy trained in one task could be used advantageously to accelerate learning in a different task; and secondly, that MIKT is a successful approach for achieving such a knowledge transfer. This works even when the teacher and student MDPs have different state- and action-spaces, and is realized by learning embeddings that are task-aligned and are optimized with a mutual information loss (Algorithm~\ref{algo:complete}).


\subsection{ABLATION STUDIES}
\textbf{Are gradients from both $\{L_{\text{PPO}}, L_{\text{MI}}\}$ to the encoder beneficial?} To quantify this, we experiment with two variants of our algorithm, each of which removes one of the components: {\em MIKT w/o MI}, which does not update $\phi$ with the mutual information loss proposed in Section~\ref{subsec:mi_reg}, and {\em MIKT w/o RL gradients}, which omits using the policy-gradient and the value function TD-error gradient for the encoder. Figure~\ref{fig:ablations} plots the performance of these variants and compares it to MIKT (which includes both the losses). We note that {\em MIKT w/o MI} generally struggles to learn in the early stages of training; see for instance Figure~\ref{fig:ablations} (c), (d). {\em MIKT w/o RL gradients} does comparatively better early on in training, but it is evident that MIKT is the most performant, both in terms of early training efficiency and the average episodic returns of the final policy. This supports our design choice of using both $\{L_{\text{PPO}}, L_{\text{MI}}\}$ to update the encoder $\phi$.

\begin{figure}[h]
  \centering
  \begin{subfigure}{.45\columnwidth}
    \centering
    \includegraphics[width=4cm, height=4cm]{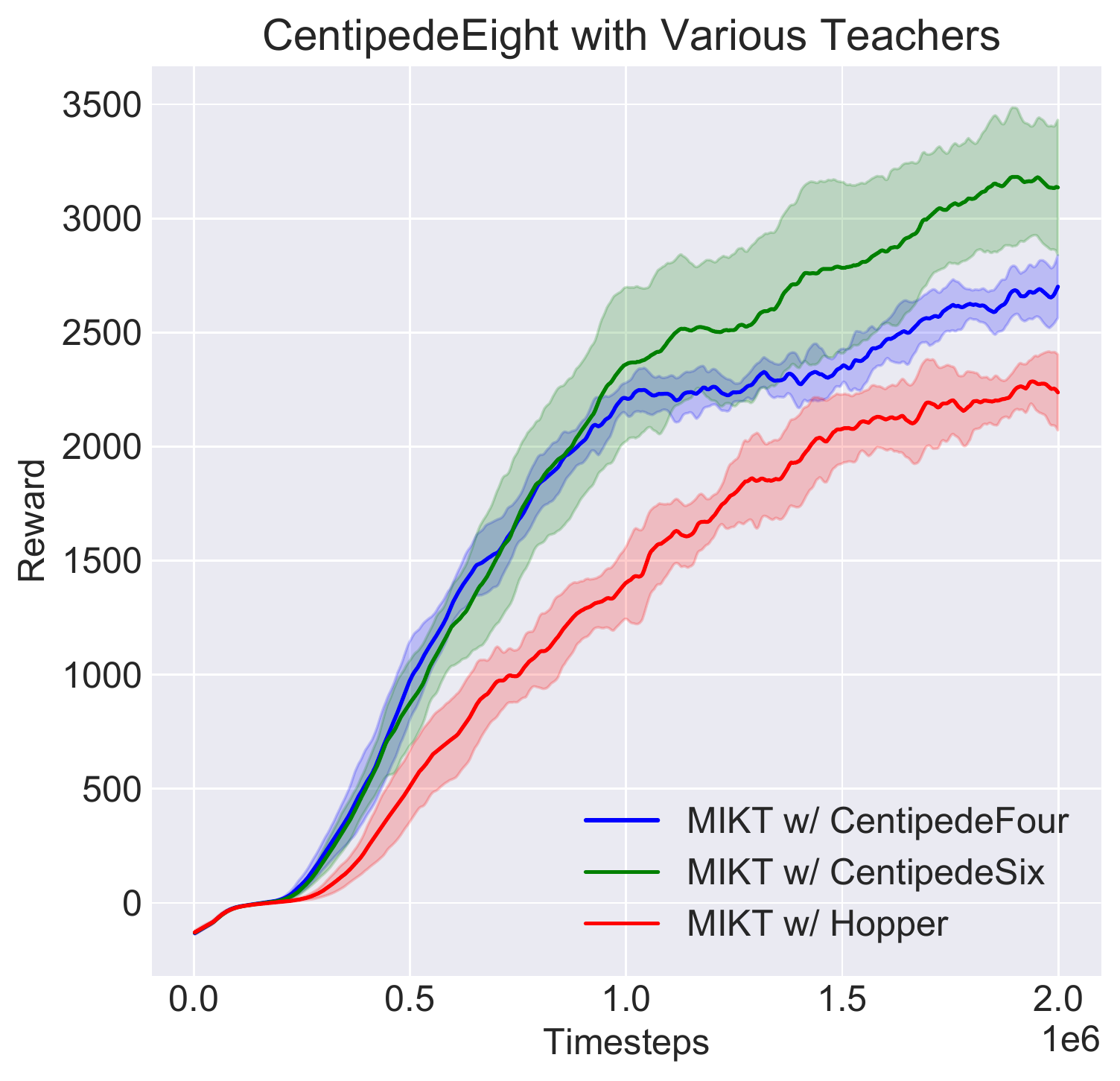}
    \caption{}
    \label{fig:abl_dissimilar_perf}
  \end{subfigure}%
  \hfill
  \begin{subfigure}{.45\columnwidth}
    \centering
    \includegraphics[width=4cm, height=4cm]{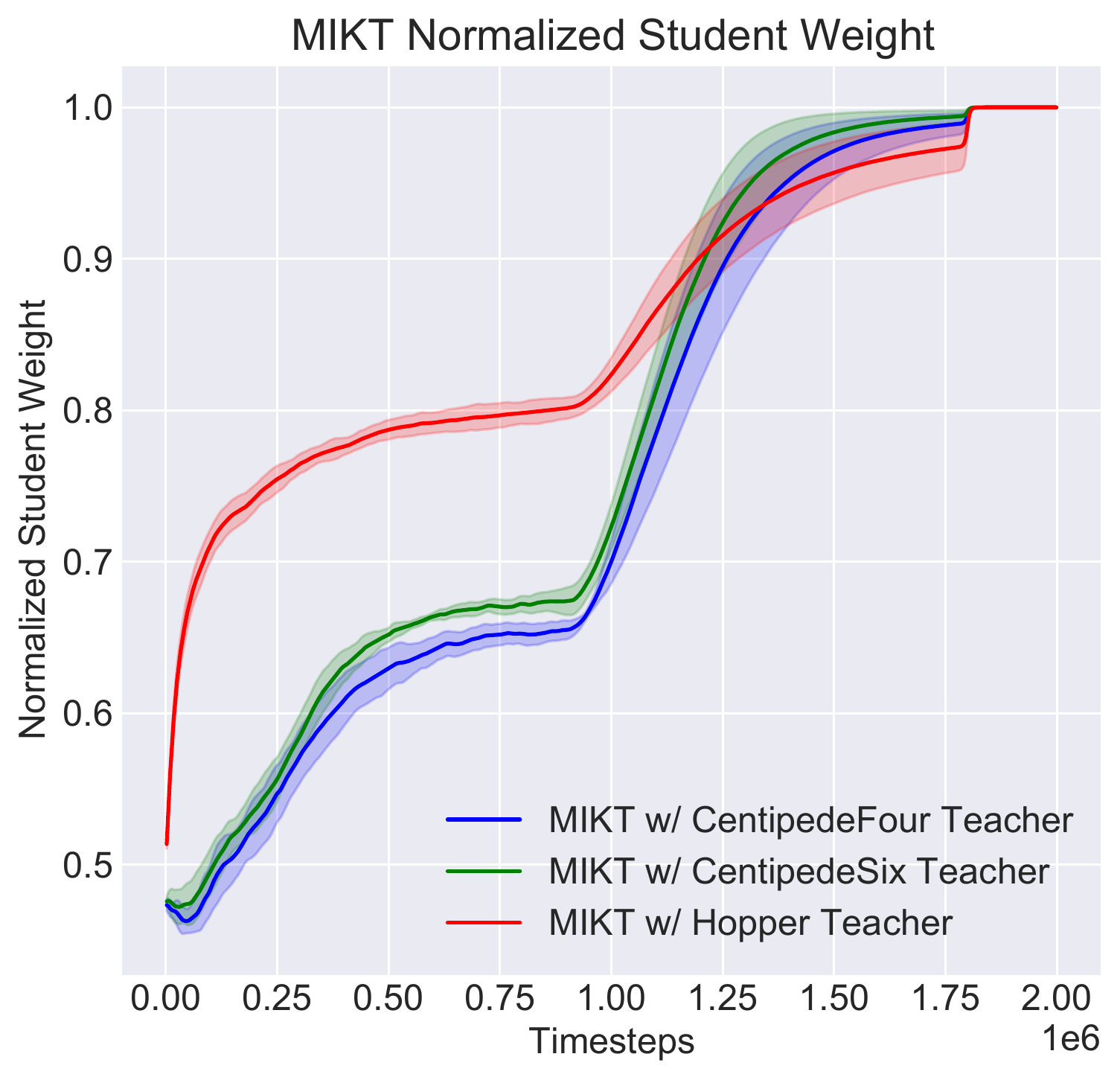}
    \caption{}
    \label{fig:abl_dissimilar_wts}
  \end{subfigure}
  \caption{Training on CentipedeEight with different teachers. (a) Transfer from a dissimilar teacher (Hopper) is less effective compared to using Centipede teachers. (b) Value of the weight on the student representation in the weighed linear combination. With the Hopper teacher, the value rises sharply in the early stages, indicating a low teacher contribution.}
\end{figure}

\textbf{How sensitive is MIKT to the task-similarity?} It is reasonable to assume that the benefits of transfer learning depend on the task-similarity between the teacher and the student. To better understand this in the context of our algorithm, we consider learning in the CentipedeEight environment using different types of teachers -- {CentipedeFour, CentipedeSix, Hopper}. In Figure~\ref{fig:abl_dissimilar_perf}, we notice that the influence of the Centipede-\{Four,Six\} teachers is much more significant than the Hopper teacher. This is likely because the motion of the centipedes shares similarity, whereas the Hopper (which is trained to hop) is a dissimilar task and therefore less useful for transfer learning. In Figure~\ref{fig:abl_dissimilar_wts} we plot the value of the weight on the student representation, when doing a weighed linear combination with the teacher (Section~\ref{subsec:task_aligned_space}). We observe with the Hopper teacher that, very early in training, the student learns to trust its own learned representations rather than incorporate knowledge from the dissimilar teacher.

%
%

\section{CONCLUSION}
In this paper, we proposed an algorithm for transfer learning in RL where the teacher (source) and the student (task) agents can have arbitrarily different state- and action-spaces. We achieve this by learning an encoder to produce embeddings that draw out useful representations from the teacher networks. We argue that training the encoder with both the RL-loss and the mutual information-loss yields rich representations; we provide empirical validation for this as well. Our experiments on a set of challenging locomotion tasks involving many-legged centipedes show that MIKT is a successful approach for achieving knowledge transfer when the teacher and student MDPs have mismatched state- and action-space.


\bibliographystyle{uai_2020}
\bibliography{uai_2020}

\begin{thebibliography}{32}
\providecommand{\natexlab}[1]{#1}
\providecommand{\url}[1]{\texttt{#1}}
\expandafter\ifx\csname urlstyle\endcsname\relax
  \providecommand{\doi}[1]{doi: #1}\else
  \providecommand{\doi}{doi: \begingroup \urlstyle{rm}\Url}\fi

\bibitem[Abbeel \& Ng(2004)Abbeel and Ng]{abbeel2004apprenticeship}
Pieter Abbeel and Andrew~Y Ng.
\newblock Apprenticeship learning via inverse reinforcement learning.
\newblock In \emph{Proceedings of the twenty-first international conference on
  Machine learning}, pp.\ ~1, 2004.

\bibitem[Agakov \& Barber(2004)Agakov and Barber]{agakov2004variational}
Felix Agakov and David Barber.
\newblock Variational information maximization for neural coding.
\newblock In \emph{International Conference on Neural Information Processing},
  pp.\  543--548. Springer, 2004.

\bibitem[Ahn et~al.(2019)Ahn, Hu, Damianou, Lawrence, and
  Dai]{ahn2019variational}
Sungsoo Ahn, Shell~Xu Hu, Andreas Damianou, Neil~D Lawrence, and Zhenwen Dai.
\newblock Variational information distillation for knowledge transfer.
\newblock In \emph{Proceedings of the IEEE Conference on Computer Vision and
  Pattern Recognition}, pp.\  9163--9171, 2019.

\bibitem[Akkaya et~al.(2019)Akkaya, Andrychowicz, Chociej, Litwin, McGrew,
  Petron, Paino, Plappert, Powell, Ribas, et~al.]{akkaya2019solving}
Ilge Akkaya, Marcin Andrychowicz, Maciek Chociej, Mateusz Litwin, Bob McGrew,
  Arthur Petron, Alex Paino, Matthias Plappert, Glenn Powell, Raphael Ribas,
  et~al.
\newblock Solving rubik's cube with a robot hand.
\newblock \emph{arXiv preprint arXiv:1910.07113}, 2019.

\bibitem[Brockman et~al.(2016)Brockman, Cheung, Pettersson, Schneider,
  Schulman, Tang, and Zaremba]{gymbrockman}
Greg Brockman, Vicki Cheung, Ludwig Pettersson, Jonas Schneider, John Schulman,
  Jie Tang, and Wojciech Zaremba.
\newblock Openai gym, 2016.

\bibitem[Czarnecki et~al.(2018)Czarnecki, Jayakumar, Jaderberg, Hasenclever,
  Teh, Osindero, Heess, and Pascanu]{czarnecki2018mix}
Wojciech~Marian Czarnecki, Siddhant~M Jayakumar, Max Jaderberg, Leonard
  Hasenclever, Yee~Whye Teh, Simon Osindero, Nicolas Heess, and Razvan Pascanu.
\newblock Mix\&match-agent curricula for reinforcement learning.
\newblock \emph{arXiv preprint arXiv:1806.01780}, 2018.

\bibitem[Czarnecki et~al.(2019)Czarnecki, Pascanu, Osindero, Jayakumar,
  Swirszcz, and Jaderberg]{czarnecki2019distilling}
Wojciech~Marian Czarnecki, Razvan Pascanu, Simon Osindero, Siddhant~M
  Jayakumar, Grzegorz Swirszcz, and Max Jaderberg.
\newblock Distilling policy distillation.
\newblock \emph{arXiv preprint arXiv:1902.02186}, 2019.

\bibitem[Duan et~al.(2016)Duan, Schulman, Chen, Bartlett, Sutskever, and
  Abbeel]{duan2016rl}
Yan Duan, John Schulman, Xi~Chen, Peter~L Bartlett, Ilya Sutskever, and Pieter
  Abbeel.
\newblock Rl2: Fast reinforcement learning via slow reinforcement learning.
\newblock \emph{arXiv preprint arXiv:1611.02779}, 2016.

\bibitem[Finn et~al.(2017)Finn, Abbeel, and Levine]{finn2017model}
Chelsea Finn, Pieter Abbeel, and Sergey Levine.
\newblock Model-agnostic meta-learning for fast adaptation of deep networks.
\newblock In \emph{Proceedings of the 34th International Conference on Machine
  Learning-Volume 70}, pp.\  1126--1135. JMLR. org, 2017.

\bibitem[Gamrian \& Goldberg(2018)Gamrian and Goldberg]{gamrian2018transfer}
Shani Gamrian and Yoav Goldberg.
\newblock Transfer learning for related reinforcement learning tasks via
  image-to-image translation.
\newblock \emph{arXiv preprint arXiv:1806.07377}, 2018.

\bibitem[Gangwani \& Peng(2017)Gangwani and Peng]{gangwani2017policy}
Tanmay Gangwani and Jian Peng.
\newblock Policy optimization by genetic distillation.
\newblock \emph{arXiv preprint arXiv:1711.01012}, 2017.

\bibitem[Hinton et~al.(2015)Hinton, Vinyals, and Dean]{hinton2015distilling}
Geoffrey Hinton, Oriol Vinyals, and Jeff Dean.
\newblock Distilling the knowledge in a neural network.
\newblock \emph{arXiv preprint arXiv:1503.02531}, 2015.

\bibitem[Lillicrap et~al.(2015)Lillicrap, Hunt, Pritzel, Heess, Erez, Tassa,
  Silver, and Wierstra]{lillicrap2015continuous}
Timothy~P Lillicrap, Jonathan~J Hunt, Alexander Pritzel, Nicolas Heess, Tom
  Erez, Yuval Tassa, David Silver, and Daan Wierstra.
\newblock Continuous control with deep reinforcement learning.
\newblock \emph{arXiv preprint arXiv:1509.02971}, 2015.

\bibitem[Liu et~al.(2019)Liu, Peng, and Schwing]{liu2019knowledge}
Iou-Jen Liu, Jian Peng, and Alexander~G Schwing.
\newblock Knowledge flow: Improve upon your teachers.
\newblock \emph{arXiv preprint arXiv:1904.05878}, 2019.

\bibitem[Mnih et~al.(2015)Mnih, Kavukcuoglu, Silver, Rusu, Veness, Bellemare,
  Graves, Riedmiller, Fidjeland, Ostrovski, et~al.]{mnih2015human}
Volodymyr Mnih, Koray Kavukcuoglu, David Silver, Andrei~A Rusu, Joel Veness,
  Marc~G Bellemare, Alex Graves, Martin Riedmiller, Andreas~K Fidjeland, Georg
  Ostrovski, et~al.
\newblock Human-level control through deep reinforcement learning.
\newblock \emph{Nature}, 518\penalty0 (7540):\penalty0 529--533, 2015.

\bibitem[Mnih et~al.(2016)Mnih, Badia, Mirza, Graves, Lillicrap, Harley,
  Silver, and Kavukcuoglu]{mnih2016asynchronous}
Volodymyr Mnih, Adria~Puigdomenech Badia, Mehdi Mirza, Alex Graves, Timothy
  Lillicrap, Tim Harley, David Silver, and Koray Kavukcuoglu.
\newblock Asynchronous methods for deep reinforcement learning.
\newblock In \emph{International Conference on Machine Learning}, pp.\
  1928--1937, 2016.

\bibitem[Ng et~al.(2000)Ng, Russell, et~al.]{ng2000algorithms}
Andrew~Y Ng, Stuart~J Russell, et~al.
\newblock Algorithms for inverse reinforcement learning.
\newblock In \emph{Icml}, pp.\  663--670, 2000.

\bibitem[Parisotto et~al.(2015)Parisotto, Ba, and
  Salakhutdinov]{parisotto2015actor}
Emilio Parisotto, Jimmy~Lei Ba, and Ruslan Salakhutdinov.
\newblock Actor-mimic: Deep multitask and transfer reinforcement learning.
\newblock \emph{arXiv preprint arXiv:1511.06342}, 2015.

\bibitem[Rajeswaran et~al.(2017)Rajeswaran, Kumar, Gupta, Vezzani, Schulman,
  Todorov, and Levine]{rajeswaran2017learning}
Aravind Rajeswaran, Vikash Kumar, Abhishek Gupta, Giulia Vezzani, John
  Schulman, Emanuel Todorov, and Sergey Levine.
\newblock Learning complex dexterous manipulation with deep reinforcement
  learning and demonstrations.
\newblock \emph{arXiv preprint arXiv:1709.10087}, 2017.

\bibitem[Rezende \& Mohamed(2015)Rezende and Mohamed]{rezende2015variational}
Danilo~Jimenez Rezende and Shakir Mohamed.
\newblock Variational inference with normalizing flows.
\newblock \emph{arXiv preprint arXiv:1505.05770}, 2015.

\bibitem[Rozantsev et~al.(2018)Rozantsev, Salzmann, and
  Fua]{rozantsev2018beyond}
Artem Rozantsev, Mathieu Salzmann, and Pascal Fua.
\newblock Beyond sharing weights for deep domain adaptation.
\newblock \emph{IEEE transactions on pattern analysis and machine
  intelligence}, 41\penalty0 (4):\penalty0 801--814, 2018.

\bibitem[Rusu et~al.(2015)Rusu, Colmenarejo, Gulcehre, Desjardins, Kirkpatrick,
  Pascanu, Mnih, Kavukcuoglu, and Hadsell]{rusu2015policy}
Andrei~A Rusu, Sergio~Gomez Colmenarejo, Caglar Gulcehre, Guillaume Desjardins,
  James Kirkpatrick, Razvan Pascanu, Volodymyr Mnih, Koray Kavukcuoglu, and
  Raia Hadsell.
\newblock Policy distillation.
\newblock \emph{arXiv preprint arXiv:1511.06295}, 2015.

\bibitem[Rusu et~al.(2016)Rusu, Rabinowitz, Desjardins, Soyer, Kirkpatrick,
  Kavukcuoglu, Pascanu, and Hadsell]{rusu2016progressive}
Andrei~A Rusu, Neil~C Rabinowitz, Guillaume Desjardins, Hubert Soyer, James
  Kirkpatrick, Koray Kavukcuoglu, Razvan Pascanu, and Raia Hadsell.
\newblock Progressive neural networks.
\newblock \emph{arXiv preprint arXiv:1606.04671}, 2016.

\bibitem[Schmitt et~al.(2018)Schmitt, Hudson, Zidek, Osindero, Doersch,
  Czarnecki, Leibo, Kuttler, Zisserman, Simonyan,
  et~al.]{schmitt2018kickstarting}
Simon Schmitt, Jonathan~J Hudson, Augustin Zidek, Simon Osindero, Carl Doersch,
  Wojciech~M Czarnecki, Joel~Z Leibo, Heinrich Kuttler, Andrew Zisserman, Karen
  Simonyan, et~al.
\newblock Kickstarting deep reinforcement learning.
\newblock \emph{arXiv preprint arXiv:1803.03835}, 2018.

\bibitem[Schulman et~al.(2015{\natexlab{a}})Schulman, Levine, Abbeel, Jordan,
  and Moritz]{schulman2015trust}
John Schulman, Sergey Levine, Pieter Abbeel, Michael Jordan, and Philipp
  Moritz.
\newblock Trust region policy optimization.
\newblock In \emph{International Conference on Machine Learning}, pp.\
  1889--1897, 2015{\natexlab{a}}.

\bibitem[Schulman et~al.(2015{\natexlab{b}})Schulman, Moritz, Levine, Jordan,
  and Abbeel]{schulman2015high}
John Schulman, Philipp Moritz, Sergey Levine, Michael Jordan, and Pieter
  Abbeel.
\newblock High-dimensional continuous control using generalized advantage
  estimation.
\newblock \emph{arXiv preprint arXiv:1506.02438}, 2015{\natexlab{b}}.

\bibitem[Schulman et~al.(2017)Schulman, Wolski, Dhariwal, Radford, and
  Klimov]{schulman2017proximal}
John Schulman, Filip Wolski, Prafulla Dhariwal, Alec Radford, and Oleg Klimov.
\newblock Proximal policy optimization algorithms.
\newblock \emph{arXiv preprint arXiv:1707.06347}, 2017.

\bibitem[Silver et~al.(2016)Silver, Huang, Maddison, Guez, Sifre, Van
  Den~Driessche, Schrittwieser, Antonoglou, Panneershelvam, Lanctot,
  et~al.]{silver2016mastering}
David Silver, Aja Huang, Chris~J Maddison, Arthur Guez, Laurent Sifre, George
  Van Den~Driessche, Julian Schrittwieser, Ioannis Antonoglou, Veda
  Panneershelvam, Marc Lanctot, et~al.
\newblock Mastering the game of go with deep neural networks and tree search.
\newblock \emph{nature}, 529\penalty0 (7587):\penalty0 484--489, 2016.

\bibitem[Sutton et~al.(2000)Sutton, McAllester, Singh, and
  Mansour]{sutton2000policy}
Richard~S Sutton, David~A McAllester, Satinder~P Singh, and Yishay Mansour.
\newblock Policy gradient methods for reinforcement learning with function
  approximation.
\newblock In \emph{Advances in neural information processing systems}, pp.\
  1057--1063, 2000.

\bibitem[Teh et~al.(2017)Teh, Bapst, Czarnecki, Quan, Kirkpatrick, Hadsell,
  Heess, and Pascanu]{teh2017distral}
Yee Teh, Victor Bapst, Wojciech~M Czarnecki, John Quan, James Kirkpatrick, Raia
  Hadsell, Nicolas Heess, and Razvan Pascanu.
\newblock Distral: Robust multitask reinforcement learning.
\newblock In \emph{Advances in Neural Information Processing Systems}, pp.\
  4496--4506, 2017.

\bibitem[Wang et~al.(2018)Wang, Liao, Ba, and Fidler]{wang2018nervenet}
Tingwu Wang, Renjie Liao, Jimmy Ba, and Sanja Fidler.
\newblock Nervenet: Learning structured policy with graph neural networks.
\newblock 2018.

\bibitem[Ziebart et~al.(2008)Ziebart, Maas, Bagnell, and
  Dey]{ziebart2008maximum}
Brian~D Ziebart, Andrew~L Maas, J~Andrew Bagnell, and Anind~K Dey.
\newblock Maximum entropy inverse reinforcement learning.
\newblock In \emph{AAAI}, volume~8, pp.\  1433--1438. Chicago, IL, USA, 2008.

\end{thebibliography}
\newpage
\section*{APPENDIX}
\section*{A \quad Hyper-parameters}
\begin{table}[h]
\centering
\setlength\belowcaptionskip{-15pt}
\caption{\small Hyper-parameters used for all experiments.}
\begin{tabular}{c|c}
                       Hyperparameter&Value\\ \hline
                       
    Hidden Layers & $2$\\
    Hidden Units & $64$\\
    Activation & tanh\\
    Optimizer & Adam\\
    Learning Rate & $3 \text{ x } 10^{-4}$\\
    Epochs per Iteration & $10$\\
    Minibatch Size & $64$\\
    Discount $\left(\gamma\right)$ & $0.99$\\
    GAE parameter $\left(\lambda\right)$ & $0.95$\\
    Clip range $\left(\epsilon\right)$ & $0.2$\\
\end{tabular}
\label{table:hyperparameters}
\end{table}

\section*{B \quad Normalized Student Weights}

\begin{figure}[h]
\centering
\begin{minipage}{.33\textwidth}
\includegraphics[width=\textwidth, height=4cm]{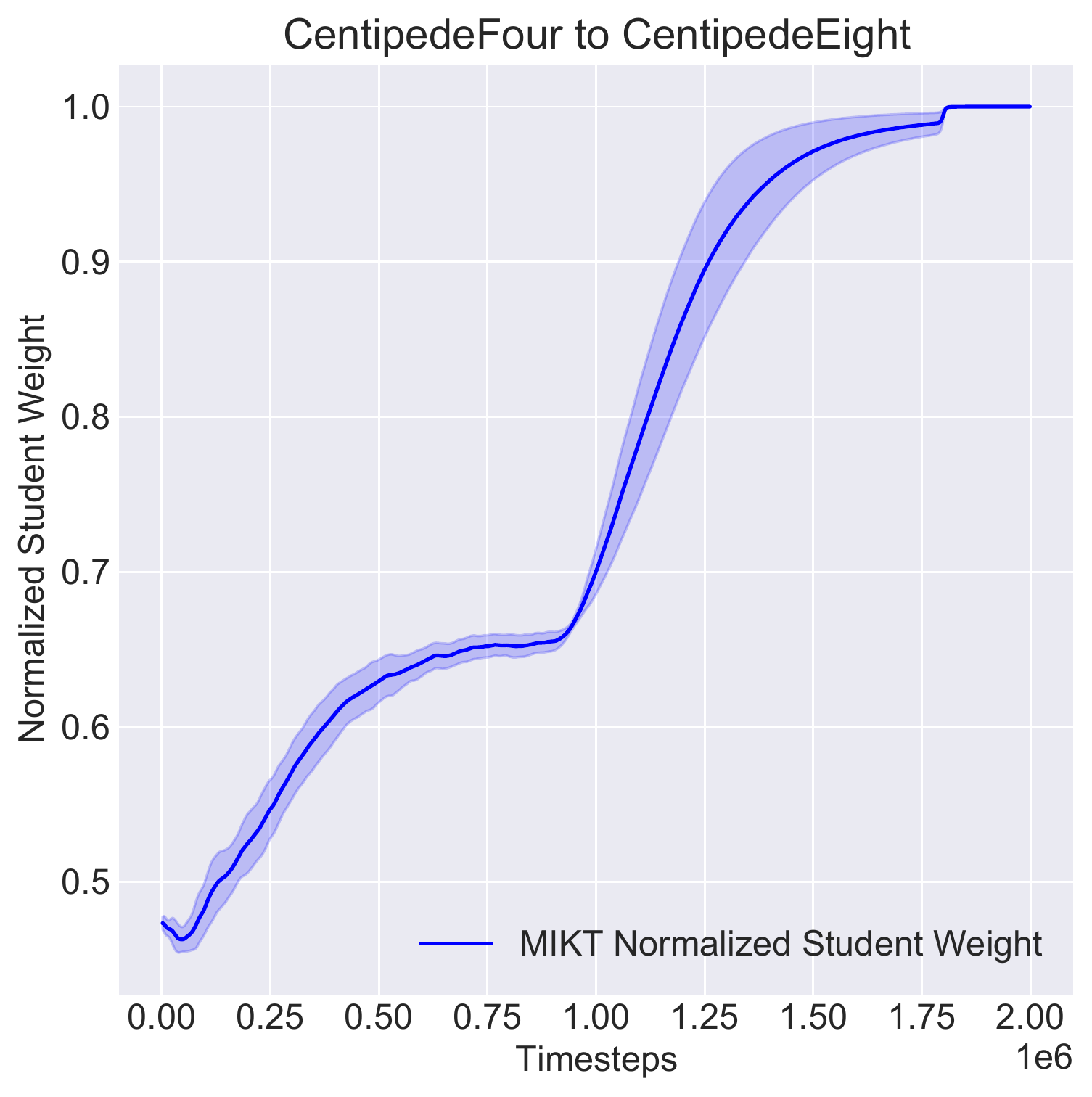} \\ \centering (a)
\end{minipage}%
\begin{minipage}{.33\textwidth}
\includegraphics[width=\textwidth, height=4cm]{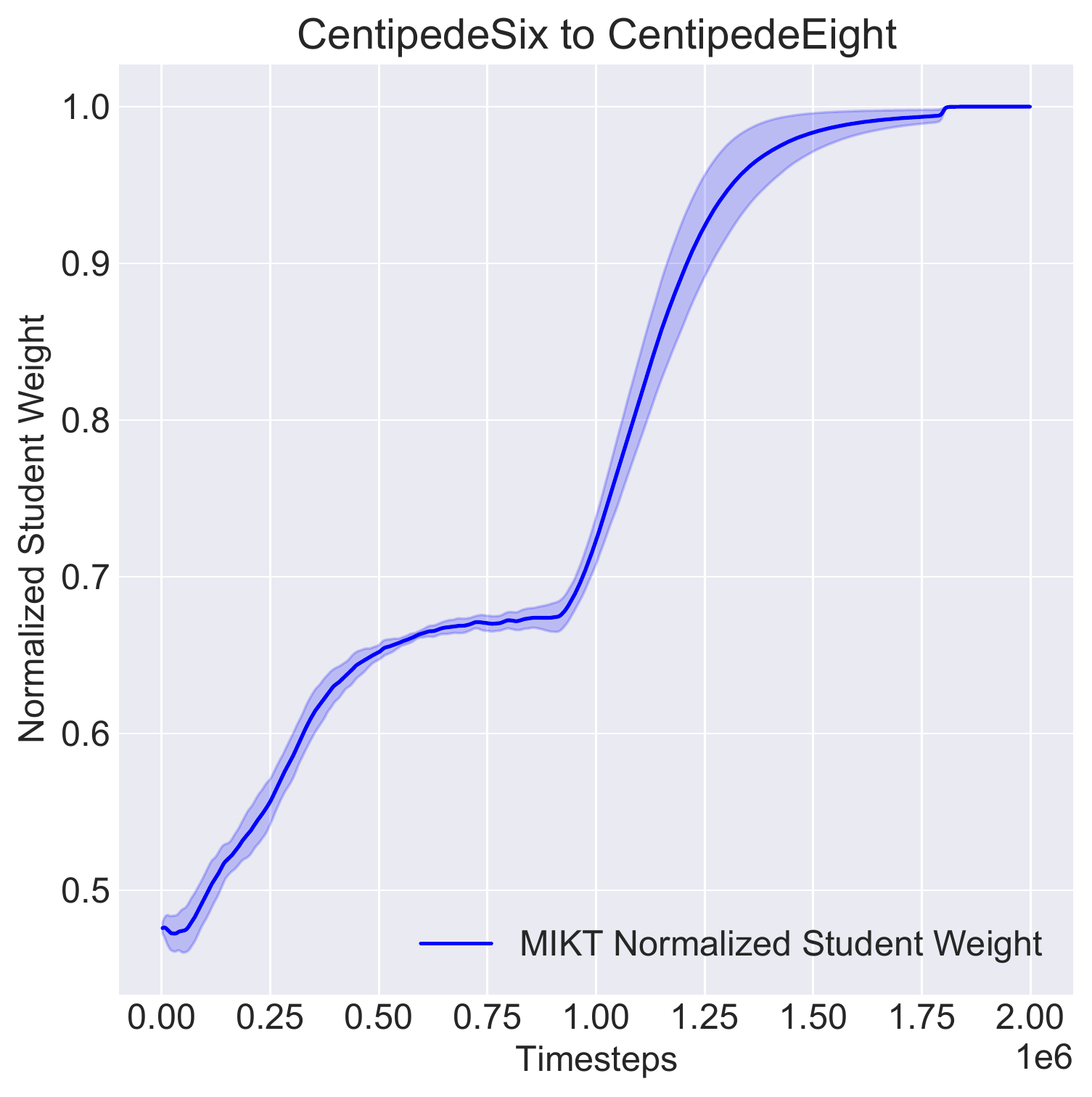} \\ \centering (b)
\end{minipage}%
\begin{minipage}{.33\textwidth}
\includegraphics[width=\textwidth, height=4cm]{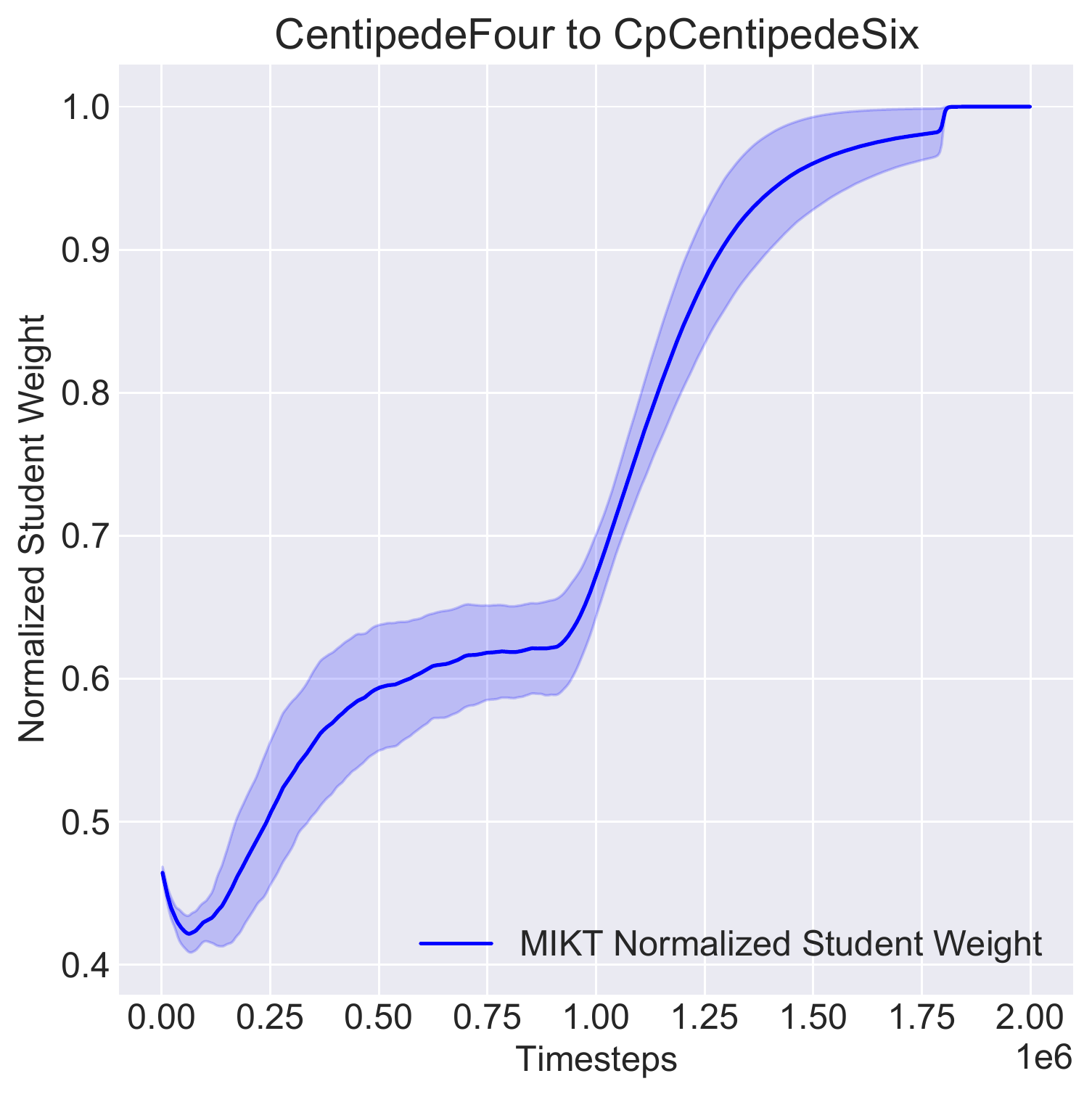} \\ \centering (c)
\end{minipage}

\medskip

\begin{minipage}{.33\textwidth}
\includegraphics[width=\textwidth, height=4cm]{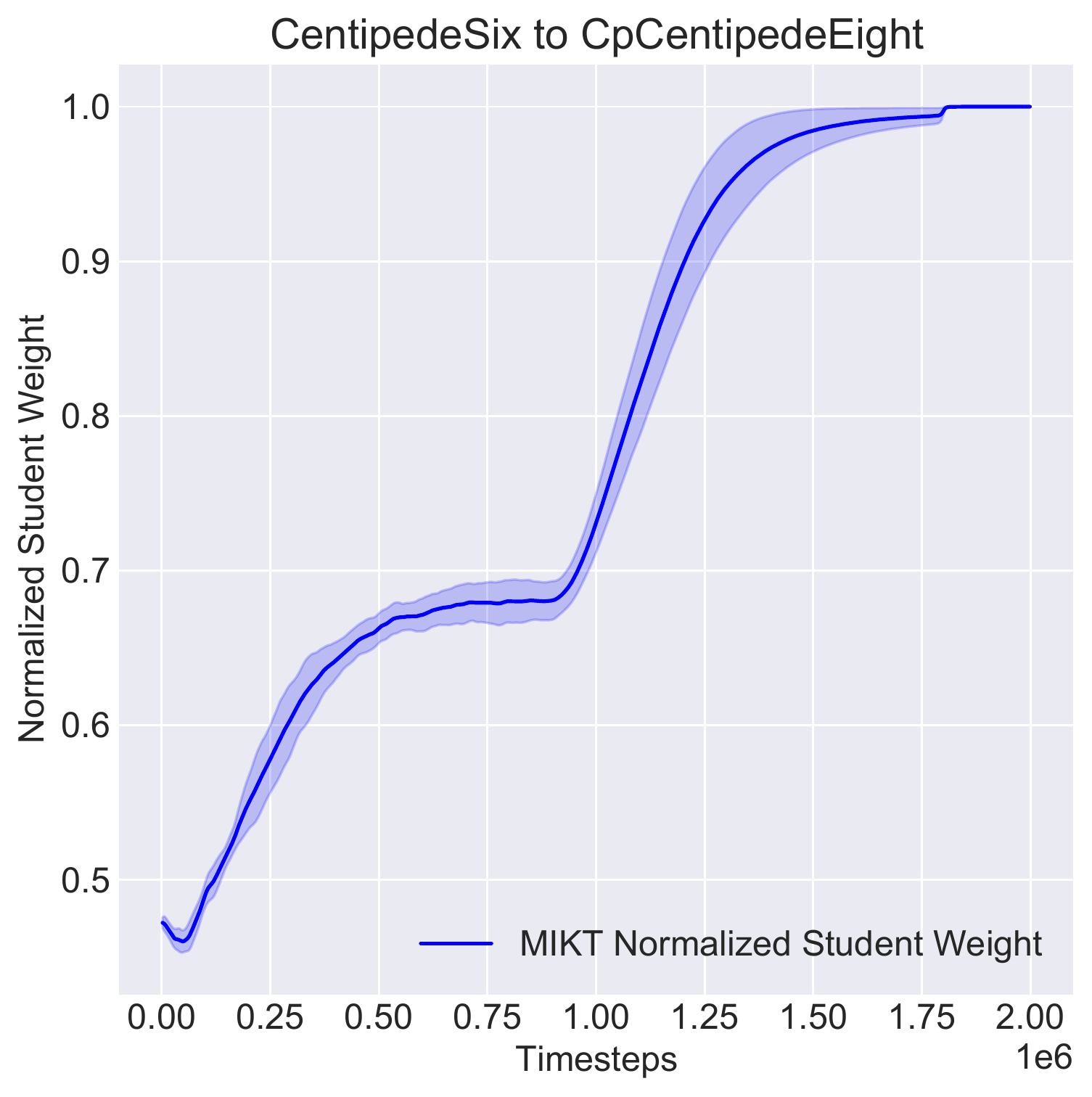} \\ \centering (d)
\end{minipage}%
\begin{minipage}{.33\textwidth}
\includegraphics[width=\textwidth, height=4cm]{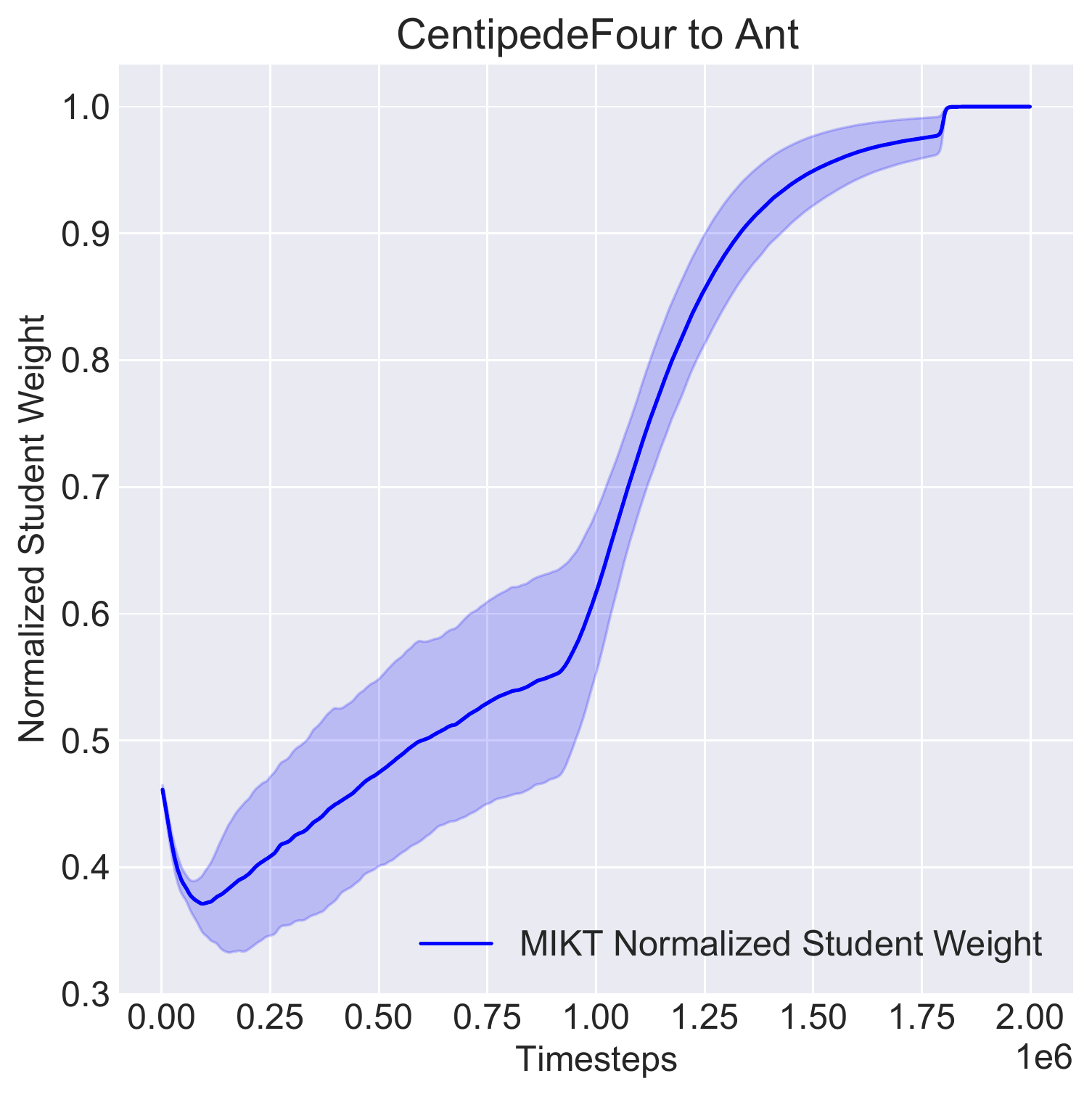} \\ \centering (e)
\end{minipage}%
\begin{minipage}{.33\textwidth}
\includegraphics[width=\textwidth, height=4cm]{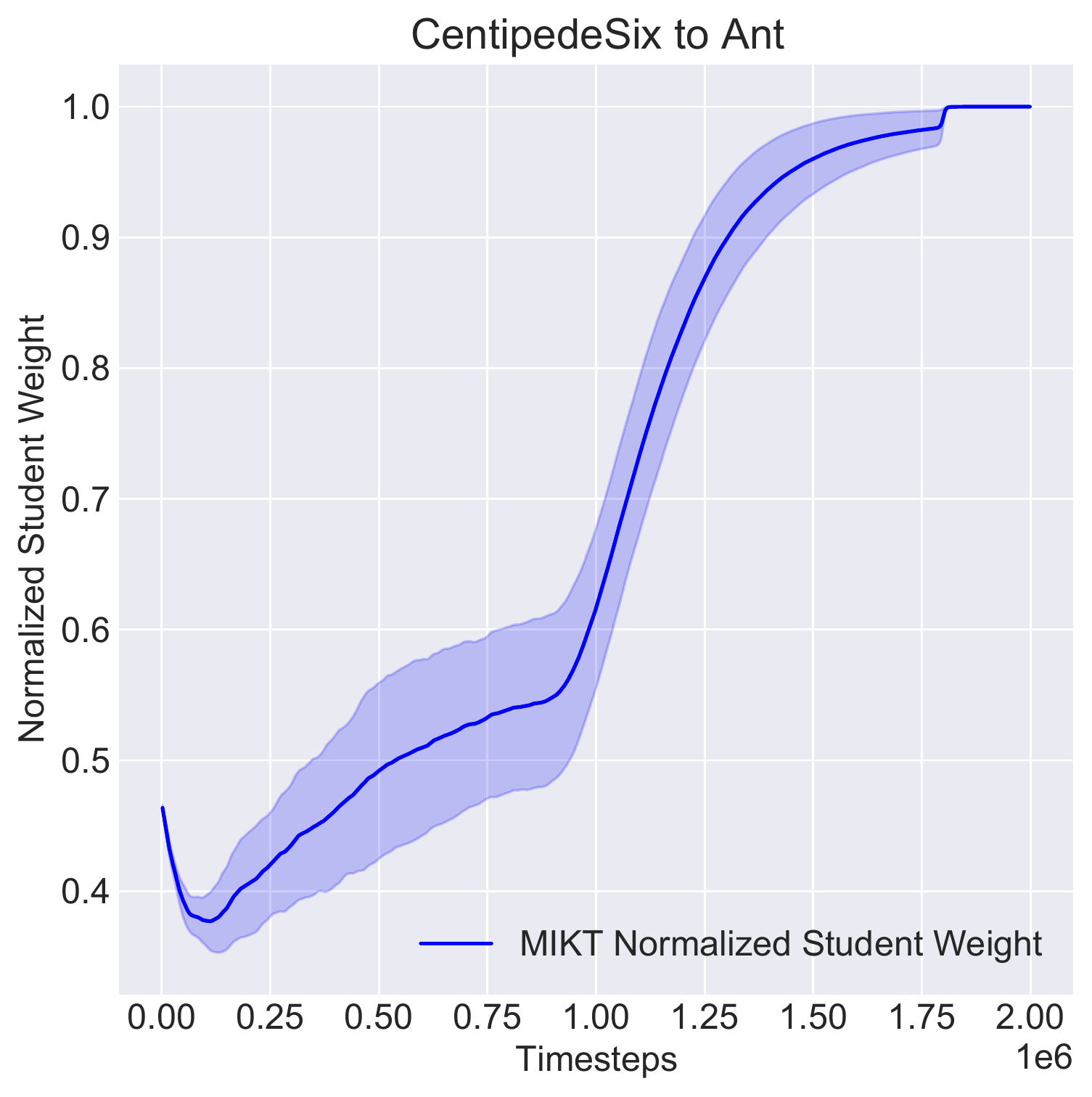} \\ \centering (f)
\end{minipage}

\medskip

\begin{minipage}{\textwidth}
\caption{Plots of the normalized student weight throughout the course of training. Lower values indicate heavier dependence on teacher representations. A value of 1 indicates the student is completely independent of the teacher.}
\end{minipage}
\label{fig:student_ps}
\end{figure}

\FloatBarrier
\clearpage

\section*{C \quad KL-Regularization Ablation}
\begin{figure}[h]
\centering
\begin{minipage}{.33\textwidth}
\includegraphics[width=\textwidth, height=4cm]{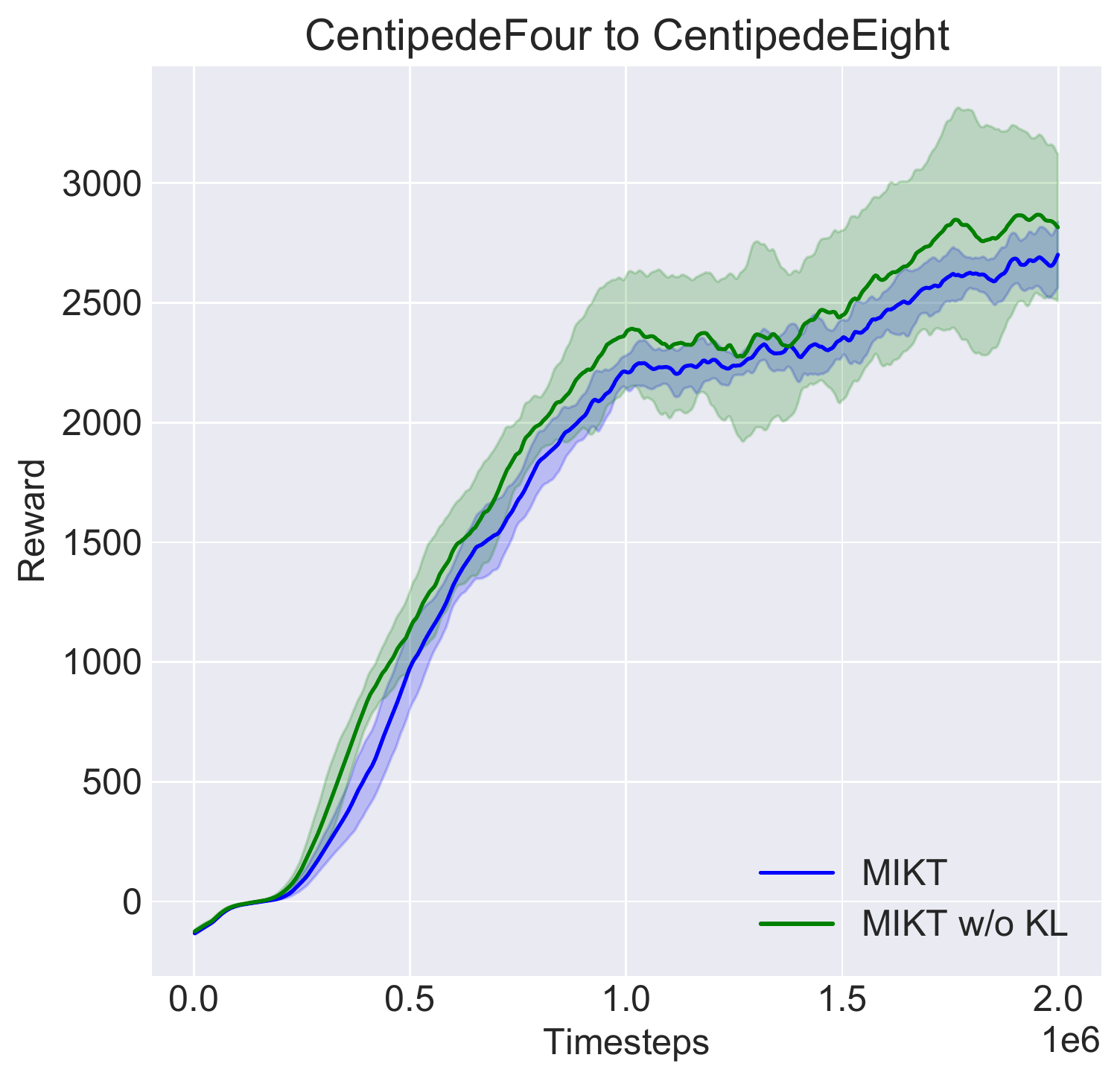} \\ \centering (a)
\end{minipage}%
\begin{minipage}{.33\textwidth}
\includegraphics[width=\textwidth, height=4cm]{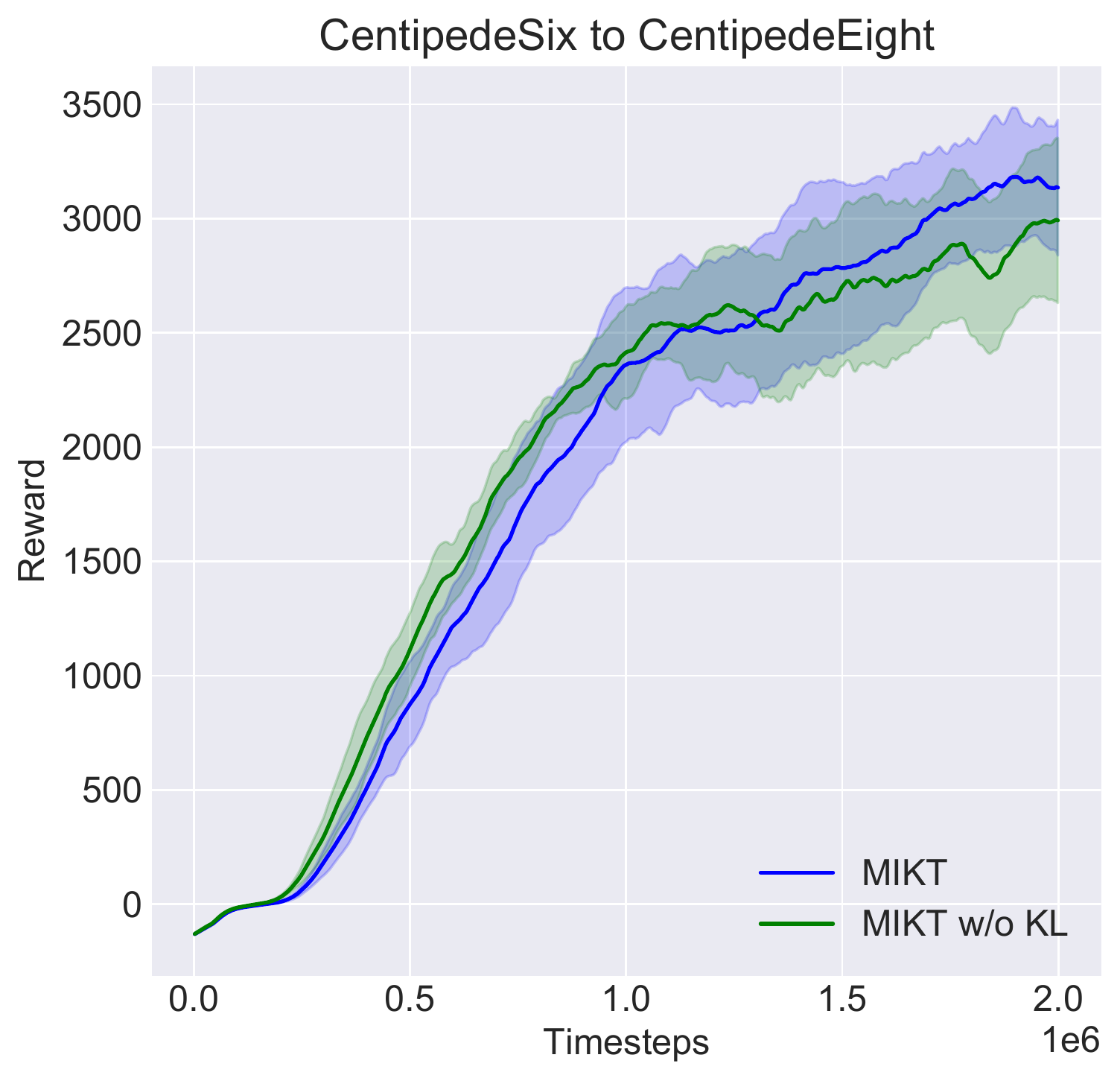} \\ \centering (b)
\end{minipage}%
\begin{minipage}{.33\textwidth}
\includegraphics[width=\textwidth, height=4cm]{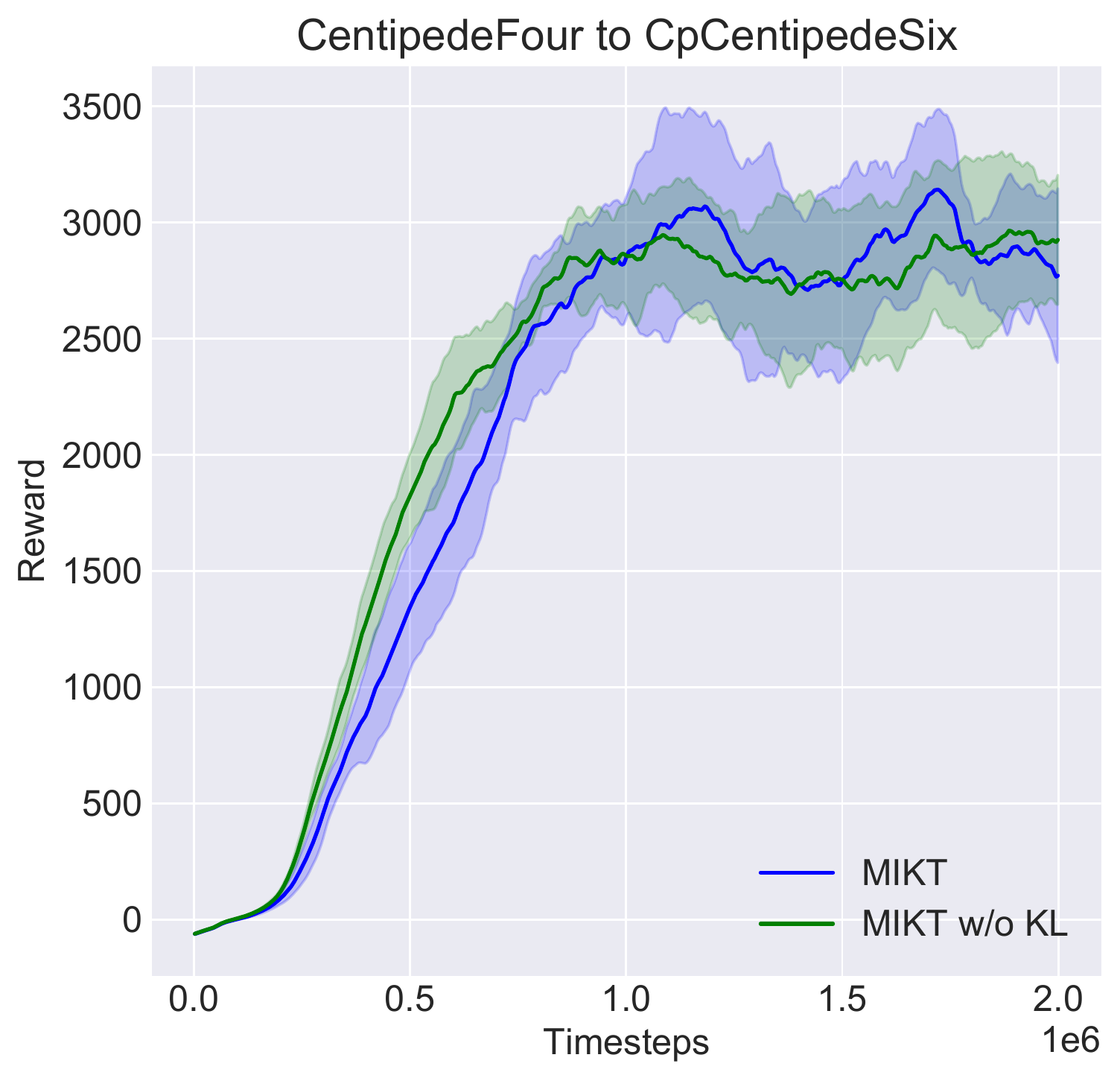} \\ \centering (c)
\end{minipage}

\medskip

\begin{minipage}{.33\textwidth}
\includegraphics[width=\textwidth, height=4cm]{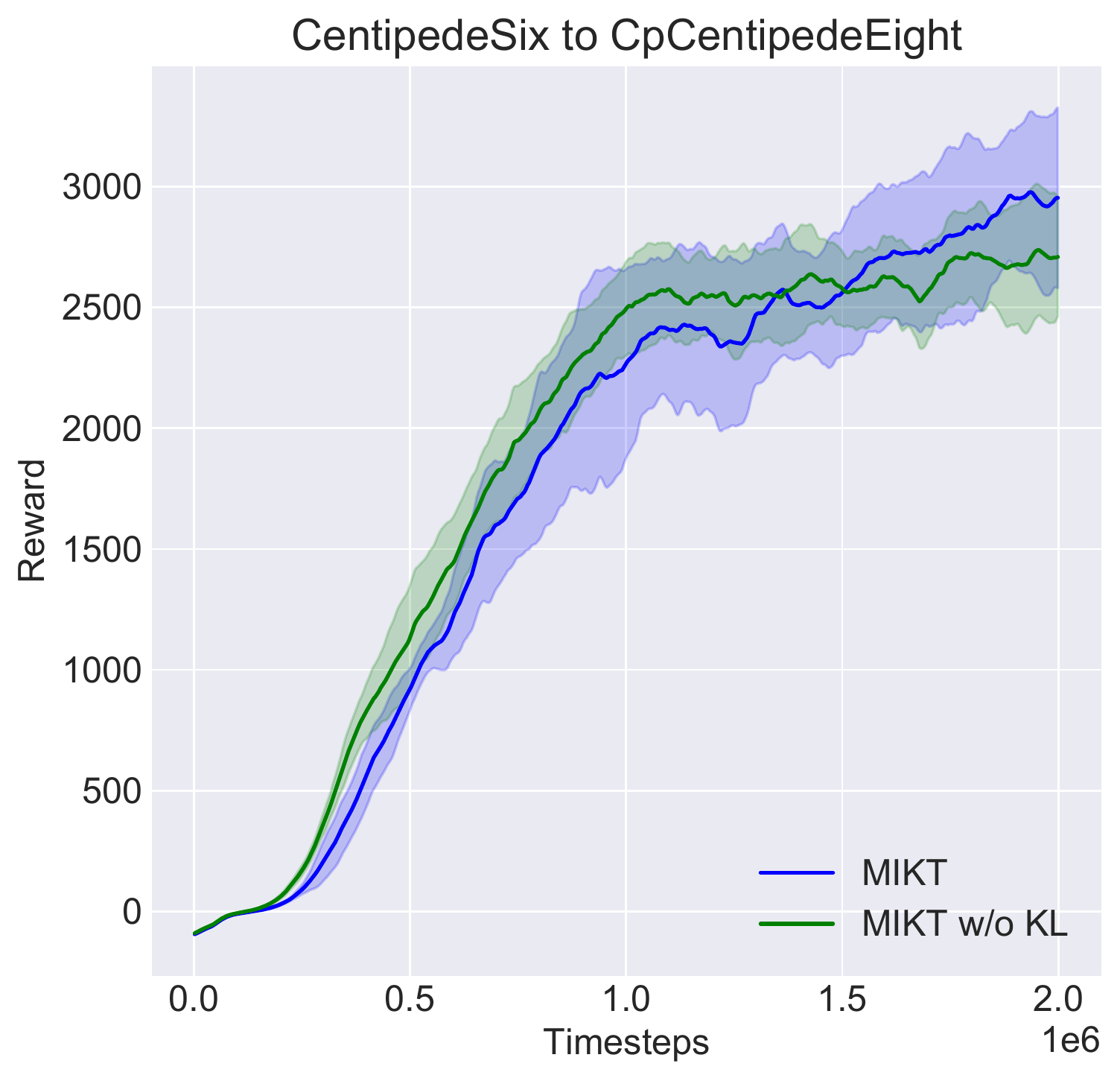} \\ \centering (d)
\end{minipage}%
\begin{minipage}{.33\textwidth}
\includegraphics[width=\textwidth, height=4cm]{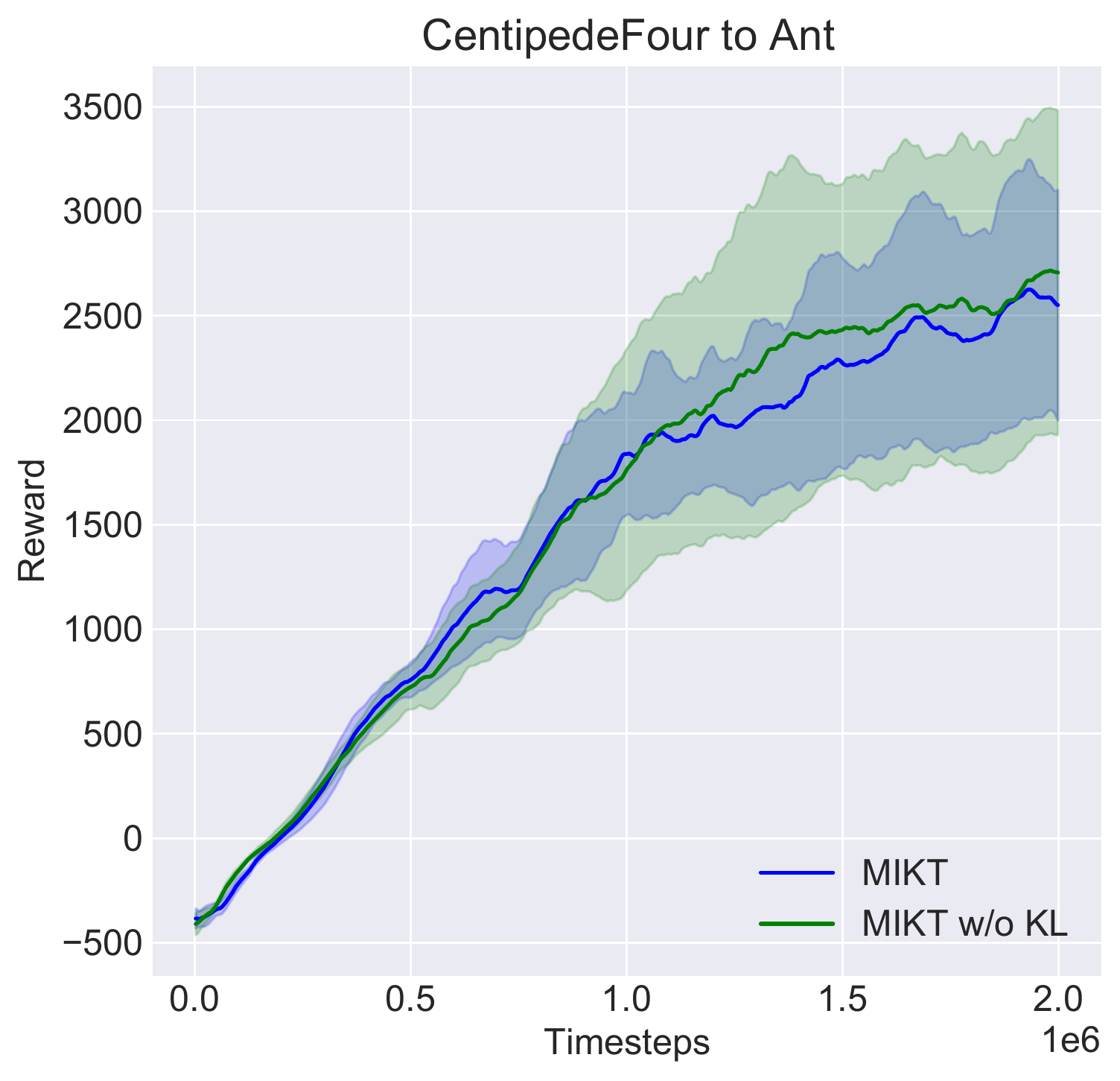} \\ \centering (e)
\end{minipage}%
\begin{minipage}{.33\textwidth}
\includegraphics[width=\textwidth, height=4cm]{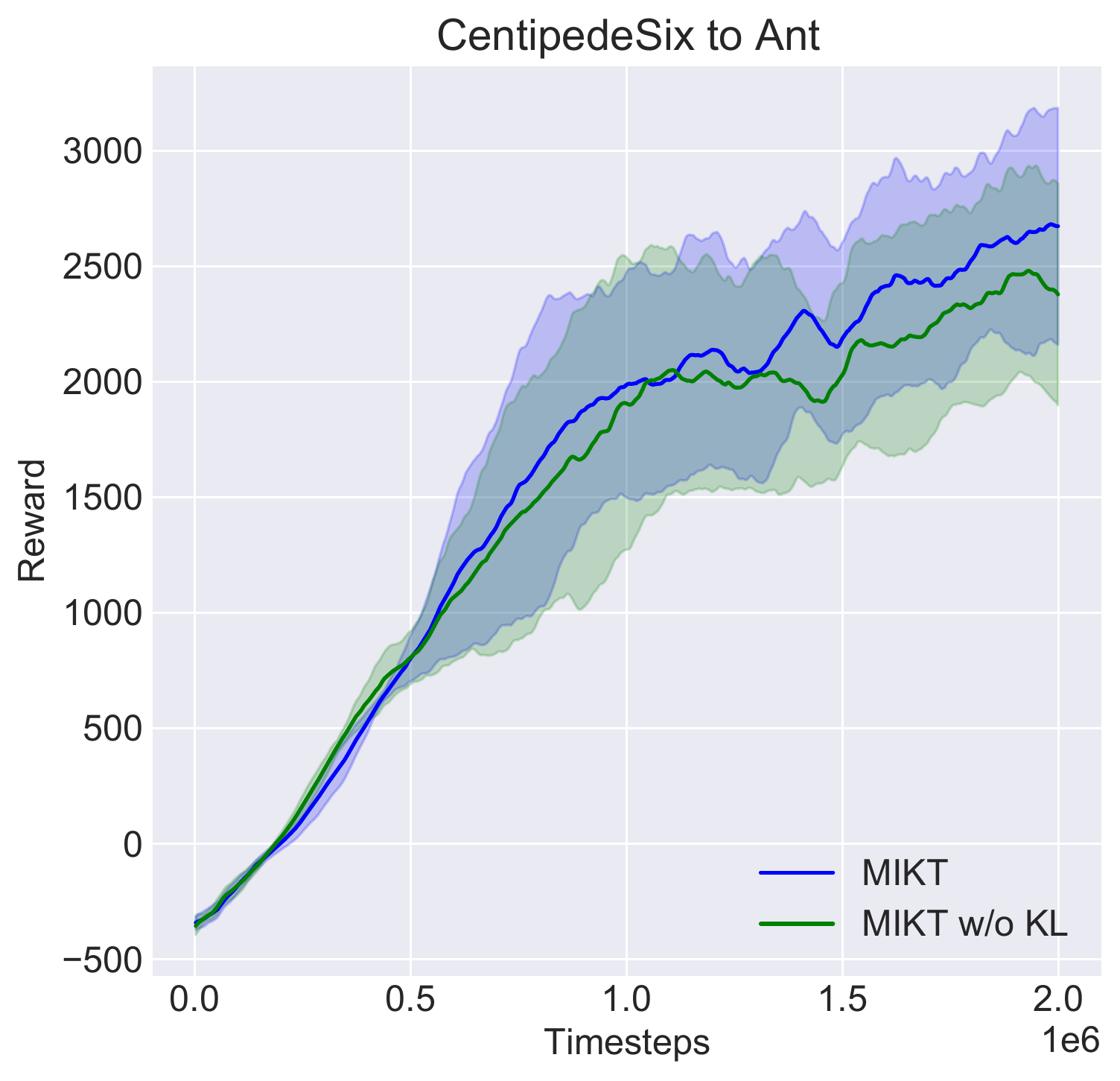} \\ \centering (f)
\end{minipage}

\medskip

\begin{minipage}{\textwidth}
\caption{MIKT with KL-regularization (blue) vs. MIKT without KL-regularization (green). MIKT still works well without the KL-regularization.}
\end{minipage}
\label{fig:kl_ablation}
\end{figure}
\end{document}